\documentclass[runningheads]{llncs}
\usepackage[T1]{fontenc}
\usepackage{graphicx}
\usepackage[english]{babel}
\usepackage{csquotes}

\usepackage{xcolor}
\usepackage{amsmath,amssymb,mathtools}
\usepackage{amsfonts}
\usepackage{siunitx}
\usepackage{verbatim}
\usepackage{ulem}
\usepackage{subcaption}
\usepackage{enumitem}
\setlist[enumerate]{leftmargin=1em, topsep=0pt}
\setlist[itemize]{leftmargin=1em, topsep=0pt}
\usepackage{xcolor}
\usepackage{booktabs,array,colortbl,tabularx}
\usepackage[export]{adjustbox}

\newcommand{\mr}[1]{\textcolor{red!60!black}{#1}}

\usepackage{hyperref}

\let\emph\textit
\newcommand{\iouresultstables}{Tables~\ref{tab:iou-per-bg-cat-voc}, \ref{tab:iou-per-bg-cat-imagenet-net2vec}, \ref{tab:iou-per-bg-cat-imagenet-gloce}}
\newcommand{\iouresultstablesimagenet}{Tables\,\ref{tab:iou-per-bg-cat-imagenet-net2vec}, \ref{tab:iou-per-bg-cat-imagenet-gloce}}
\newcommand{\iouresultstablesvoc}{\autoref{tab:iou-per-bg-cat-voc}}

\begin{document}

\title{On Background Bias of Post-Hoc Concept Embeddings in Computer Vision DNNs}

\author{Gesina Schwalbe\inst{1}\orcidID{0000-0003-2690-2478}
\and 
Georgii Mikriukov\inst{2,3}\orcidID{0000-0002-2494-6285} 
\and
Edgar Heinert\inst{5}
\and
Stavros Gerolymatos\inst{4}\orcidID{0009-0008-0610-726X} 
\and
Mert Keser\inst{3,6}\orcidID{0000-0001-7373-437X}
\and
Alois Knoll\inst{6}\orcidID{0000-0003-4840-076X}
\and
Matthias Rottmann\inst{5}\orcidID{0000-0003-3840-0184}
\and
Annika Mütze\inst{5}\orcidID{0009-0006-4155-2735} 
}
\titlerunning{Background Bias in C-XAI}
%
\authorrunning{G. Schwalbe et al.}
%
\institute{University of Lübeck, Germany \\
\email{\{firstname.lastname\}@lübeck.de}\\
\and
Hochschule Anhalt, Germany \\
\email{\{firstname.lastname\}@hs-anhalt.de}\\
\and
Continental AG, Germany\\
\email{\{firstname.lastname\}@uni-heidelberg.de}\\
\and
University of Liverpool, United Kingdom\\
\email{s.gerolymatos@liverpool.ac.uk}\\
\and
University of Wuppertal, Germany\\
\email{\{lastname\}@uni-wuppertal.de}\\
\and
Technical University of Munich, Germany\\
\email{\{mert.keser\}@tum.de}, \email{\{k\}@tum.de} \\
}

\maketitle

\begin{abstract}
    The thriving research field of concept-based explainable artificial intelligence (C-XAI) investigates how human-interpretable semantic concepts embed in the latent spaces of deep neural networks (DNNs). Post-hoc approaches therein use a set of examples to specify a concept, and determine its embeddings in DNN latent space using data driven techniques.
    This proved useful to uncover biases between different target (\emph{foreground} or concept) classes.
    However, given that the \emph{background} is mostly uncontrolled during training, an important question has been left unattended so far: Are/to what extent are state-of-the-art, data-driven post-hoc C-XAI approaches themselves prone to biases with respect to their \emph{backgrounds}?
    E.g., \textsf{wild animals} mostly occur against \textsf{vegetation} backgrounds, and they seldom appear on \textsf{roads}. Even simple and robust C-XAI methods might abuse this shortcut for enhanced performance. A dangerous performance degradation of the concept-corner cases of animals on the road could thus remain undiscovered.
    This work validates and thoroughly confirms that established Net2Vec-based concept segmentation techniques frequently capture background biases, including alarming ones, such as underperformance on road scenes.
    For the analysis, we compare 3 established techniques from the domain of background randomization on >50 concepts from 2 datasets, and 7 diverse DNN architectures.%
    \footnote{Code available at: \url{https://github.com/gesina/bg_randomized_loce}} Our results indicate that even low-cost setups can provide both valuable insight and improved background robustness.
\end{abstract}

\section{Introduction}
With the advance of deep neural networks (DNNs) into safety\mr{-}critical application areas, like medical imaging or automated driving perception, a pressing need to explain the inner workings of trained DNNs has arisen \cite{%
euaiact,
iso8800,
keser2024generative,
koopman2019ul4600
}.
To this end, an increasingly popular class of introspection methods utilizes concept-based explainable artificial intelligence (C-XAI): Their common aim is to verify and explore the internal representation of \emph{concepts} from natural language within DNN latent spaces \cite{poeta2023conceptbased,lee2024neural,schwalbe2022concept}.
For verification purposes, special focus was placed on post-hoc supervised C-XAI techniques based on TCAV~\cite{kim2018interpretability} and Net2Vec~\cite{fong2018net2vec}. These reveal, for already trained DNNs, whether and how user-defined concepts are represented in the DNN intermediate outputs \cite{%
achtibat2023attribution,
schwalbe2021verification
}.

As a shared basic principle, they use few positive and negative examples of a concept to train a simple, usually linear \cite{kim2018interpretability} model. This model, which we refer to as the \emph{concept embedding (CE)}, shall distinguish between intermediate outputs of image regions with or without the concept of interest.
Performance and similarity of CEs provides valuable insights into the semantic structure and potential biases of the DNN's latent space \cite{kim2018interpretability,olah2017feature}.
Prominent examples from the computer vision domain are gender bias (tie cooccurring with male \cite{kim2018interpretability}), dependence on object size \cite{schwalbe2021verification,mikriukov2023evaluating}, and nonconformity to symbolic rules \cite{%
giunchiglia2022roadr
}.

CEs even give rise to very intuitive interpretations: They capture the most activating neurons/filters for a concept; and represent the latent space vector pointing into the direction of \enquote{more} concept, e.g., looks more \enquote{animalish}.
However, in this work, we choose a model-based definition to highlight a crucial caveat: Supervised CEs are data-driven, and therefore prone to mirror biases inherent to the concept data and to the DNN intermediate outputs.
In particular, the C-XAI results may be sensitive to bias of the DNN-learned concepts with respect to their backgrounds, e.g., large animals rarely occur against a background of roads. If unexplored, these biases may conceal important flaws of the DNN's generalization capability, e.g., failing to detect animals crossing the road.
This would be no surprise: Recent work by Janousková \textit{et al.} \cite{Janoukov2024SegmentTR} demonstrated on the example of fungal classification, how background information can significantly influence model decision-making. Substrate features were often leveraged by models even when they were not explicitly part of the target class.
Apart from full DNNs, bias exploration of CEs has so far been restricted to bias between different foreground concepts \cite{achtibat2023attribution,schwalbe2021verification,kim2018interpretability} or adversarial concepts \cite{mikriukov2024unveiling} not foreground-vs.-background; and only few works have briefly touched upon background randomization for CEs, but not with background robustness in mind~\cite{mikriukov2023evaluating}.

Meanwhile, a rich field of research has been established around reducing background biases during the training phase of a DNN: Many techniques for background randomization, mostly via image manipulation, are available to average out any background dependencies.
Early simple ones crop and paste an image's foreground onto new backgrounds \cite{tobin2017domain}. Diverse techniques complement this, allowing for diversification of backgrounds, generation of new ones \cite{pernias2024wrstchen}, 
and enhanced realism of pasting \cite{keser2021content}. 
Even factorization of influence factors like shape, texture and color, is possible \cite{mutze2024influence}. 
In this work, we leverage a set of standard background randomization techniques to, for the first time, systematically explore and measure background biases of a broad spectrum of post-hoc supervised C-XAI methods, computer vision DNNs, and datasets.

The questions at hand are whether (1) despite being linear, CEs do fall for (natural) background biases; and (2) this can be uncovered (and avoided) using background randomization techniques, increasing chances to find otherwise hard-to-discover biases in DNN models and data. 
In this paper, we thoroughly investigate and confirm these questions for the specific but common case of concept segmentation, i.e., a region-wise classification of DNN intermediate outputs. This region-based setting makes the problem particularly handy: No strict (and costly to obtain) realism of the full image composition is needed anymore, only of spatially local patches. We therefore use and compare 3 low-cost background randomization techniques, and test these both for use in testing (for bias discovery) and training (for bias confirmation and removal) of the CEs.
This even allows to unravel sources of bias: If performance still drops for CEs trained with background randomization, this indicates a bias intrinsic to the DNN's representations.
Our main findings and contributions are as follows:
\begin{itemize}
    \item a \textbf{method and workflow to both reveal and assess background biases of concept embeddings} in DNN latent spaces, and simultaneously \textbf{counteract data-biased concept-based explanations},
    using established background randomization techniques;
    \item a broad study proving effectiveness of our method: standard DNNs and concept datasets exhibit clear and potentially dangerous background biases, which can at least partly be mitigated using local-to-global C-XAI techniques;
    \item an ablation study showing that insightful results can already be obtained very \textbf{cheaply} with as little as a \textbf{single round of background replacements and a single late DNN layer}.
\end{itemize}

\section{Related Work}

\subsection{Concept-based XAI}
Concept-based explainable AI summarizes DNN introspection techniques that associate DNN internal units like filters with hu\-man-un\-der\-stand\-able concepts from natural language. In the visual domain these can be object or subobject classes, as well as object attributes (material, texture, etc.) \cite{bau2017network}, complete scenes \cite{zhou2014learning}, and more general language synonym sets \cite{fong2018net2vec,fellbaum1998wordnet}.
First methods in the subfield associated concepts with single neurons or CNN-filters. An example for this is simple unsupervised feature visualization \cite{olah2017feature,nguyen2019understanding}.
By now, the field has grown to a wide range of problems, with a growing amount of research interest and reviews on the topic \cite{%
lee2024conceptbased,
lee2024neural,%
poeta2023conceptbased,%
schwalbe2022concept,%
keser2023interpretable%
}, as detailed in the following. One branch focuses on ante-hoc enforcement of alignment between intermediate units and given concepts \cite{%
koh2020concept,
sawada2022concept
} respectively automatically learned interpretable prototypes \cite{%
chen2019this,
feifel2021reevaluating,
willard2024this
} (see, e.g., \cite[Fig.\,4]{lee2024conceptbased} for a graphical overview). These, however, require control over the training process and special care to avoid leakage of non-concept information into the units' meanings \cite{%
hoffmann2021this,%
marconato2022glancenets%
}. In many applications, post-hoc analysis of already trained feature spaces poses an alternative. While a plethora of methods by now allow to explore learned concepts of models in an unsupervised fashion \cite{%
posada-moreno2024eclad,%
vielhaben2023multidimensional,
fel2023craft,
zhang2021invertible,
ghorbani2019automatic
}, verification typically relies on finding specific, user-predefined concepts from given rules \cite{%
achtibat2023attribution,
giunchiglia2022roadr,
schwalbe2021verification
}.
Good results were achieved using complex latent-space-to-concept mappings, like neuralizing flows \cite{esser2020disentangling} 
or non-linear SVM classifiers \cite{crabbe2022concept}.
However, simple linear mappings are claimed to be more intuitive for explanations \cite{kim2018interpretability}. 

An early representative of this direction is the assignment of concepts to (singular) filters by Bau et al.~\cite{bau2017network}. The problem setting later evolved to more general assignment of concepts to latent space \emph{vectors}. This was first done in a supervised manner by Kim et al.~in TCAV \cite{kim2018interpretability} for concept classification, and in parallel for concept segmentation by Fong et al.~as the Net2Vec framework \cite{fong2018net2vec}, and later regression \cite{graziani2018regression}. Here, we extend this line of concept segmentation research. To this end, we build on several proposed extensions of this framework, including less costly preprocessing \cite{schwalbe2021verification,mikriukov2023evaluating}, and more stable losses \cite{schwalbe2022enabling}. Most recently, Mikriukov et al.\ introduced a local-to-global version of Net2Vec, LoCE \cite{mikriukov2024local}, to analyze how concepts manifest differently across individual samples. This additionally captures the variance in concept representations and supposedly their dependence on surrounding context, such as background information.

\subsection{Background Randomization}
To randomize image backgrounds, one typically first identifies the foreground canvas. This process often relies on existing segmentation labels, but can also involve AI-based approaches, such as foundational segmentation models \cite{Picek2024AnimalIW,kirillov2023segment} or foreground-prediction models \cite{Simoni2022UnsupervisedOL,you2025focusuniversalforegroundsegmentation}. 

A straightforward approach for background randomization is to paste the identified foreground onto a different background image. This technique has been used for background augmentation \cite{ryali2021characterizing}, to analyze the influence of foreground and background information on classification models \cite{xiao2021noise}, and to introduce out-of-distribution objects into semantic segmentation street scenes \cite{blum2021fishyscapes}. As progress has been made in the field of image generation \cite{rombach2022high,lugmayr2022repaint}, prompt-based generation of artificial backgrounds has been considered \cite{li2024simple,lynch2023spawrious}. These methods come with the limitations of the generative models, which include the generation of undesired foreground objects within complex backgrounds.
Another approach leaves the natural domain and corrupts backgrounds either with simple distortions, such as Gaussian noise \cite{Moayeri2022ACS} or by style transfering the background with paintings, thereby changing the texture and color of backgrounds \cite{theodoridis2022trapped}. Finally, some datasets contain naturally occurring adversarial samples, where unusual foreground-background combinations challenge model robustness \cite{hendrycks2021nae,lau2021natural,Janoukov2024SegmentTR,chan2021segmentmeifyoucan}. However, these datasets are not appropriate for a controlled study of specific foreground-background relationships, and the adversarial combinations they contain are not random.

To study the randomization of naturally occurring backgrounds and assess the individual limitations of each technique, we apply both straightforward pasting and artificial background generation independently. Furthermore, to increase background variability within a single image, we experiment with placing foregrounds on Voronoi-patch-based backgrounds, where each cell contains a different background, following the approach of \cite{mutze2024influence}.

\section{Background: Global and Local Post-hoc Concept Segmentation}\label{sec:background-ce}

Post-hoc supervised C-XAI aims to find a latent representation for a user-defined language concept inside an already trained DNN. In particular, the motivating question is: \textit{Does the DNN latent space encode (enough) information about a concept?}
Here, we use two main and closely related methods: Net2Vec and local concept embeddings (LoCE).

\subsubsection{Global Concept Embeddings.}
Starting with Net2Vec, the main ingredients in the beginning are a trained DNN, a selected layer therein, and a labeled image dataset defining the concept via segmentation masks. They now rely on three tricks to answer the above question
(cf.\ illustration in \autoref{fig:ce}):
\begin{figure}[t]
    \centering
    \vspace*{-\baselineskip}
    \includegraphics[width=.9\linewidth]{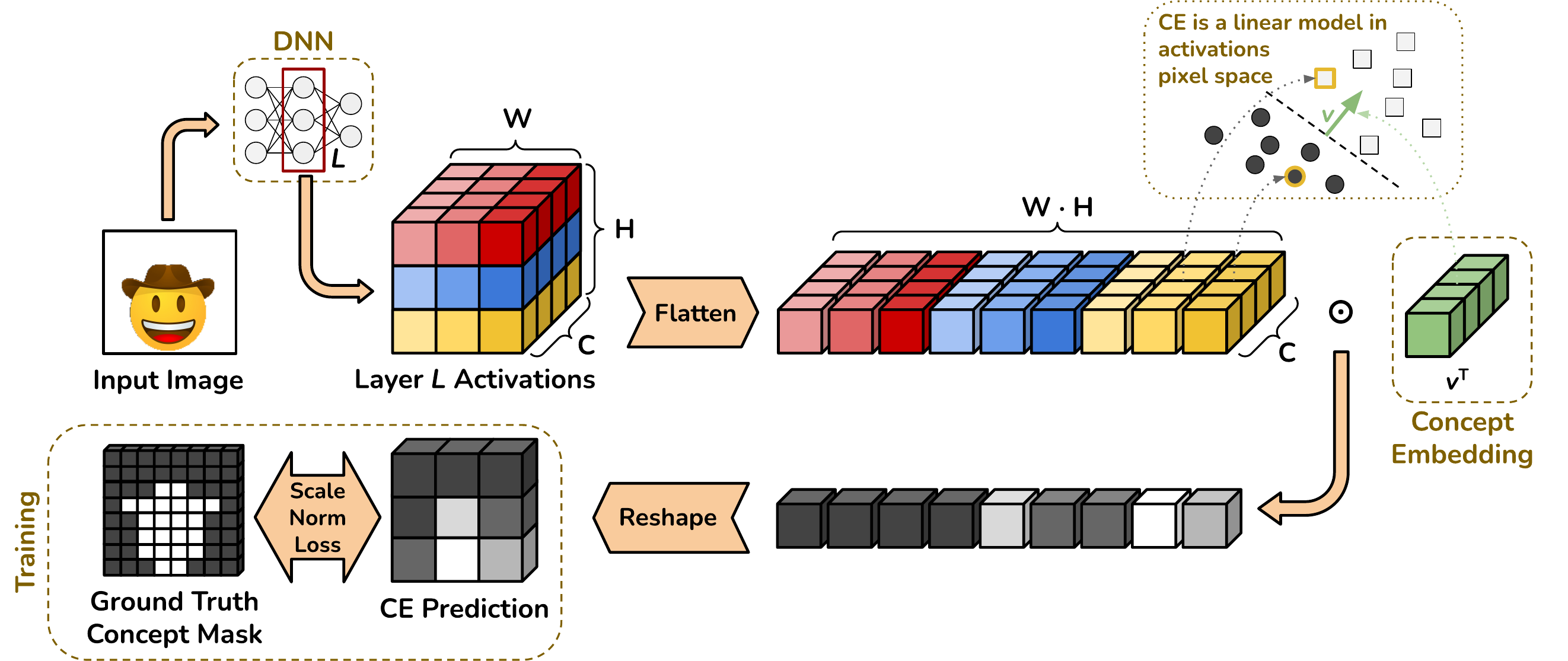}
    \caption{Illustration of training and inference of global and local concept embeddings (here for concept \textsf{cowboy smiley}): A CE is a linear classifier on activation map pixels, represented by its normal vector $v$ in latent space. To infer the model on a new \textit{input image}, (1) the \textit{activations} for the desired layer $L$ are collected, (2) each pixel is classified using dot product with the CE, which yields a heatmap of concept presence as \textit{CE prediction}. For training, (3) this is compared against a \textit{ground truth concept mask} to update the CE weights via gradient descent.
    }
    \label{fig:ce}
\end{figure}
\begin{enumerate}
    \item The question is reformulated to: Is it possible to predict the presence of my concept in an image solely from the latent space outputs of the image at that layer? If this is the case, then the latent space does encode information about the concept (the mutual information is high).
    This makes it essentially a \textbf{learning problem}: Can one train a well-performing classifier to predict (segment) presence of the concept?
    \item The input image already contains enough information about the concept, since human labelers can decide its presence/segmentation. If the DNN would therefore be the identity function, the previous point would in principle still confirm that the DNN has \enquote{knowledge} about the concept.
    To counteract this counterintuitive situation, a strict \textbf{model bias} must be imposed. The choice here is a \textbf{linear model}, due to its good interpretability.
    \item Lastly, to make it a \textbf{segmentation instead of a classification task}, Net2Vec suggested making the \textbf{classification per activation map pixel} instead of the complete activation map.
    Given an activation map of size $\text{Channels}\times\text{Height}\times\text{Width}$ thus yields $\text{Height}\times\text{Width}$ activation map pixels, each a vector of dimensionality $\text{Channels}\eqqcolon C$.
    As we now extract concepts from pixels instead of whole feature maps, the linear classification problem is drastically simplified: It becomes the training of the kernel of a singular $1\times1$ convolution.
\end{enumerate}
This kernel takes the role of the normal vector to the linear hyperplane separating concept from non-concept activation map pixels in that pixels space. In particular, it is itself also a vector of size $C$ (the concept vector); the one pointing in the direction of the concept pixels.
This makes the vectors easy to compare.

\subsubsection{Measuring CE Performance.}
Note that this formulation makes obtaining CEs simply a machine learning problem: The classifier for a concept (with the concept's CE as parameters) can be trained on a training set; and later be used for inference on new samples. In our case new samples are activation map pixels of new images. So, a CE can be used to segment the learned concept in a new image. Such obtained segmentation masks can be tested against a ground truth after rescaling, measuring how well the CE captured information about the concept.

The segmentation test performance of a CE depends both on the quality of the concept training data used to obtain the CE, as well as the quality of concept encoding of the DNN latent space. We unravel this here to distinguish between background bias originating from
(1) the concept training data (removable by randomizing the concept training data's backgrounds), and 
(2) from the DNN itself (not removable).

\subsubsection{Local and Globalized Local Concept Embeddings.}
The main adaptation done by the LoCE approach \cite{mikriukov2024local} compared to Net2Vec is quite simple: Instead of training one CE on the \emph{complete} concept dataset, one trains one CE per \emph{single sample} in the concept dataset (the \emph{local CE, LoCE}, of that image). A single LoCE thus captures the information about how the occurrences of its concept in \emph{its specific image} can be differentiated from the background in \emph{that image}; as compared to a Net2Vec CE, which captures how to differentiate \emph{any} concept instance from \emph{any} background.
The distribution of the LoCEs in latent space captures as much of the variance of the separation problem as possible.
LoCEs can later be combined again via averaging, again obtaining a CE that is valid for a complete concept, not only instances from one image. We here call such an average the \emph{globalized LoCE} (GloCE) of the concept.
Mikriukov et al.\ \cite{mikriukov2024local} showed that this is similar but in general not equal to the Net2Vec (i.e., directly global) CE trained on the same concept data. Therefore, we investigate them separately (\emph{and later in this work show that they better capture background biases}).

\subsubsection{Layer Selection.}
It is known that the quality of information about a given concept typically smoothly evolves across several layers, gradually increasing until an optimum (set of) layer(s) \cite{%
fuchs2018neural,
fong2018net2vec,
rabold2020expressive
}. Also, colors and textures are typically better embedded in early layers, whereas more complex concepts are optimally embedded in later layers \cite{fong2018net2vec}.
Similar to \cite{mikriukov2024local}, this work adopts a structured approach to layer selection by extracting activations at the level of full network blocks rather than after individual within-block layers. Blocks act as meaningful processing units, providing consistent checkpoints for analyzing internal representations. For CNNs, selected convolutional block outputs are used; for transformers, encoder outputs are analyzed. In multi-branch architectures (e.g., object detectors), one representative branch is selected to trace feature progression.


\section{Approach}\label{sec:approach}

Concept embeddings are obtained using (concept) training data, and therefore can themselves easily be prone to biases.
In this paper, we are specifically interested in \emph{background biases}. These are \textit{any dependencies of the CE's performance on non-object-level features that are unrelated to the quality of the CE's concept};
e.g., a fox should still be a fox independent of whether it is on the road or on a field. Non-object-level feature here refers to a feature not part of the concept objects, i.e., one of the features in input image pixels that do not belong to the image's semantic canvas. \textit{Such features are commonly also referred to as scenery or background \cite{zhou2014learning}.}
The primary use of CEs is also what makes their biases so interesting: By design, they represent a piece of knowledge encoded in the internal representations of their DNN under scrutiny. If a CE is biased,
e.g., better detects foxes in fields than on streets,
this might indicate a flaw in the DNN's encodings. In that case, background information leaks into evidence collection for the concept \textsf{fox}.

In this work, we specifically investigate the following questions:
\begin{enumerate}
    \item Given trained CEs, \textit{does a change of the background distribution impact their performance when \emph{testing}?}
    In other words: Are there preferred or hard-to-deal-with backgrounds?
    \item \textit{Does the \emph{CE training} extract different information about the concept if the background distribution is changed during training?}
    Which divides into:
    \begin{itemize}
        \item \textit{Do different (here: uniformly randomized) backgrounds lead to \emph{different concept vector representations}?}
        This quantifies the background bias of the CE, both coming from the concept data or the DNN itself.
        \item \textit{Do CEs trained on randomized backgrounds exhibit a different / better \emph{performance} distribution over backgrounds?}
        An improvement indicates removable background bias originating from the concept training data. Remaining performance issues can indicate a bias of the DNN itself. 
    \end{itemize}
\end{enumerate}
At the heart of answering them are two main ingredients: Being able to control the background distribution of test and training datasets; and the actual training and evaluation of the CEs (cf.\ \autoref{sec:background-ce}).
In the following, we first detail our approaches to control the background distribution via background randomization techniques, and then summarize the used CE techniques and metrics.

\subsection{Techniques for Background Randomization}

\subsubsection{Image-level Background Randomization.}
In order to extract background-debiased and thus broader CE class distributions, we randomized the backgrounds of each foreground in three ways, always preserving the original canvas specified by the foreground class mask.

\subsubsection{Foreground.}
All methods have in common that the foregrounds of the concepts of interest must be given, i.e., a dataset of images together with segmentation masks defining which parts of an image belong to which of the foreground concepts.
For this we employed two well-known concept segmentation datasets as a foreground dataset:
The Pascal VOC dataset \cite{everingham2010pascal} with 20 classes, and the
ImageNetS50 dataset \cite{gao2022luss}, a subset of ImageNet1k \cite{deng2009imagenet} with 50 classes that has foreground class masks for each of its 64431 train and 752 validation images.

\subsubsection{Background Annotations \& Filtering.}
Another preliminary for our methods is that a dataset of labeled backgrounds must be available, such that the distribution of backgrounds can be controlled.
Assuming both a foreground and a set of background samples are given, we randomly select a background, resize and crop both images to size $256\times256$. We further employ two filtering criteria: We exclude background images for which the foreground class itself was in the top $5$ predictions of an ImageNet pretrained Vision Transformer (\texttt{vit\_base\_patch16\_224} \cite{dosovitskiy2020image,rw2019timm}); and for generation of several background variants we draw background classes without replacement, such that a highly diverse set of background variants is created.  
The pasting is also common to all methods: We created additional images by replacing all non-foreground pixels with random realistic scenes from the backgrounds.

\subsubsection{Background Generation.}
We here investigate three different techniques for generating the background.
As the simplest version, we directly use a 10\% subset of the Places205 dataset~\cite{zhou2014learning}, a large-scale scene collection of 205 categories, containing 247k images (\texttt{Places}). For the testing, we manually selected and clustered 24 background categories into 10 diverse, each visually homogeneous supercategories (for details see \autoref{tab:places-supclasses}).

\begin{table}[t]
    \scriptsize
    \centering
    \vspace*{-\baselineskip}
    \caption{Tested background supercategories with corresponding Places205 classes.
    Superclasses were selected to be visually homogeneous.
    \scriptsize\\\emph{ Note: crowded background sceneries were excluded to not interfere with the \textsf{person} concept class.}
    }
    \resizebox{.9\linewidth}{!}{%
    \begin{tabular}{>{\ttfamily}l >{\ttfamily}l}\toprule
        architecture:   &   abbey, aqueduct, arch, attic, basilica, building\_facade, office\_building\\\midrule
        indoors:   &   bedroom, dining\_room, hotel\_room, kitchen, kitchenette, living\_room\\\midrule
        at\_water:   &   bayou, canyon, coast, creek, dock, islet, marsh, ocean, pond\\\midrule
        machinery:   &   engine\_room\\\midrule
        open\_lands:   &   badlands, butte\\\midrule
        forest:   &   bamboo\_forest, forest\_path, rainforest\\\midrule
        botanical:   &   botanical\_garden, cottage\_garden, formal\_garden, orchard, topiary\_garden\\\midrule
        field:   &   golf\_course, wheat\_field, fairway\\\midrule
        snow:   &   crevasse, iceberg, mountain\_snowy, ski\_slope, snowfield \\\midrule
        road:   &   crosswalk, highway \\\bottomrule
    \end{tabular}
    }\vspace*{-\baselineskip}
    \label{tab:places-supclasses}
\end{table}

In an attempt to increase the variability of background randomization per image and thus reduce the amount of required samples, we used a Voronoi-patching approach similar to \cite{mutze2024influence} (\texttt{Voronoi}). In this work, we generate a Voronoi diagram \cite{torquato2002CellRandomField} based on $8$ uniformly sampled points (cf.\ \autoref{fig:voronoi} for illustration). Each cell is filled with a randomly shifted cutout either from a chosen or a randomly sampled background image.

Lastly, we investigate whether the need for manual background labeling can be overcome by using generative AI. We leverage Würstchen\footnote{\scriptsize Implementation: \url{https://huggingface.co/warp-ai/wuerstchen}, using default parameters.}~\cite{pernias2024wrstchen}, a highly efficient text-to-image diffusion model, to create semantically rich backgrounds. A diverse set of background categories---including \texttt{cloudscape}, \texttt{space}, \texttt{jungle}, \texttt{desert}, \texttt{arctic}, \texttt{volcanic}, \texttt{ocean}, and \texttt{abstract patterns} -- was defined, with each category described using detailed text prompts emphasizing realism and atmospheric characteristics. The generator synthesized $100$ high-resolution ($1024 \times 1024$) images per category using controlled diffusion, ensuring diverse and contextually relevant textures.

\subsection{Measuring background robustness}

In order to benchmark and compare the CEs, we employ two main perspectives: A model-based one relying on black-box performance measurement; and a representation-based one, measuring similarity of the underlying concept vectors.

\subsubsection{Performance Measurement.}
With respect to the model-based perspective, a CE can be simply considered as a linear classifier that can predict a segmentation mask on new samples. In these terms, robustness translates into robust performance, i.e., generalization to novel kinds of samples such as the ones with randomized backgrounds. To measure the quality of an output mask $M$ against the ground truth $M_{\text{gt}}$, we use Intersection over Union (IoU), also known as the Jaccard index, defined as
\begin{gather}
    \SwapAboveDisplaySkip
    \text{IoU}(M, M_\text{gt}) \coloneqq \frac{M\cap M_\text{gt}}{M\cup M_\text{gt}} \in [0,1]
\end{gather}
We note that measuring performance on unseen samples does not make sense for LoCEs: These are not trained to generalize. Instead, we globalize them: Given a set of LoCEs that represent one concept in a latent space, we define their globalized LoCE (GloCE) as the model / hyperplane defined by the mean of their representing concept vectors.

\subsubsection{Representation Comparison.}
As an alternative to black-box test statistics, we use one of the core advantages of linear models: Being represented by vectors, we compare CEs $c_1, c_2$ pairwise by calculating the cosine similarity of their underlying concept vectors $v_1, v_2$.
With $\|\cdot\|$ denoting the Euclidean norm, this is defined as
\begin{gather}
    \SwapAboveDisplaySkip
    \text{CosSim}(v_{1}, v_{2}) \coloneqq \frac{v_{1}^T v_{2}}{\|v_{1}\|\cdot \|v_{2}\|} \in [-1, 1]
\end{gather}
and intuitively measures the angle between the two vectors, resulting in -1 for opposite, 0 for orthogonal, and 1 for parallel vectors.
This can be used to directly compare a pair of LoCEs that was trained on the same image.


\section{Experiments}
\begin{figure}
\begin{subfigure}{\linewidth}
\caption{\textbf{Examples} of an original ImageNetS image (\textit{left}) with random Places205 background (\textit{center left}), a Voronoi-style background (\textit{center right}) and an image with Würstchen-generated background (\textit{right}):}
\vspace*{-\baselineskip}%
\label{fig:voronoi}
\begin{center}
  \begin{minipage}{.95\linewidth}%
  \hfill%
  \begin{minipage}{0.23\textwidth}
    \centering
    \includegraphics[width=\linewidth]{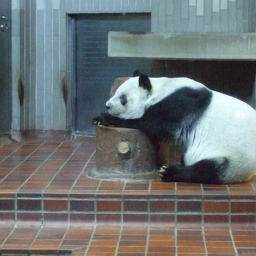}
  \end{minipage}\hfill%
  \begin{minipage}{0.23\textwidth}
    \centering
    \includegraphics[width=\linewidth]{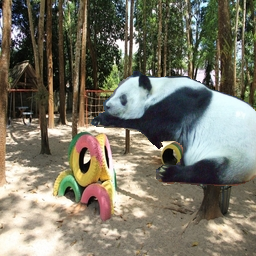}
  \end{minipage}\hfill%
  \begin{minipage}{0.23\textwidth}
    \centering
    \includegraphics[width=\linewidth]{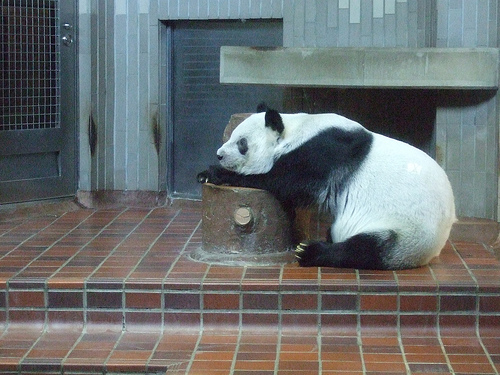}
  \end{minipage}\hfill%
  \begin{minipage}{0.23\textwidth}
    \centering
    \includegraphics[width=\linewidth]{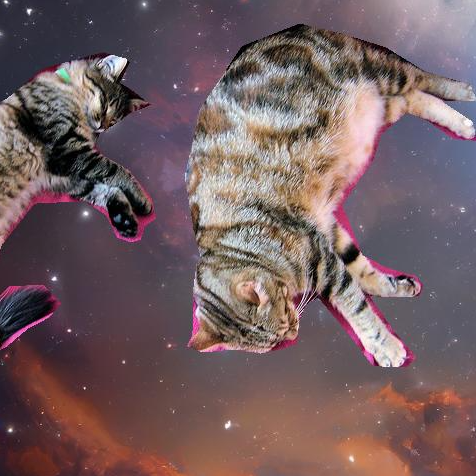}
  \end{minipage}\hfill%
  \end{minipage}%
\end{center}
\end{subfigure}
    \centering
    \begin{subfigure}{\linewidth}
        \caption{Exemplary inference results of \textbf{Net2Vec} CEs.}
        \includegraphics[width=\linewidth]{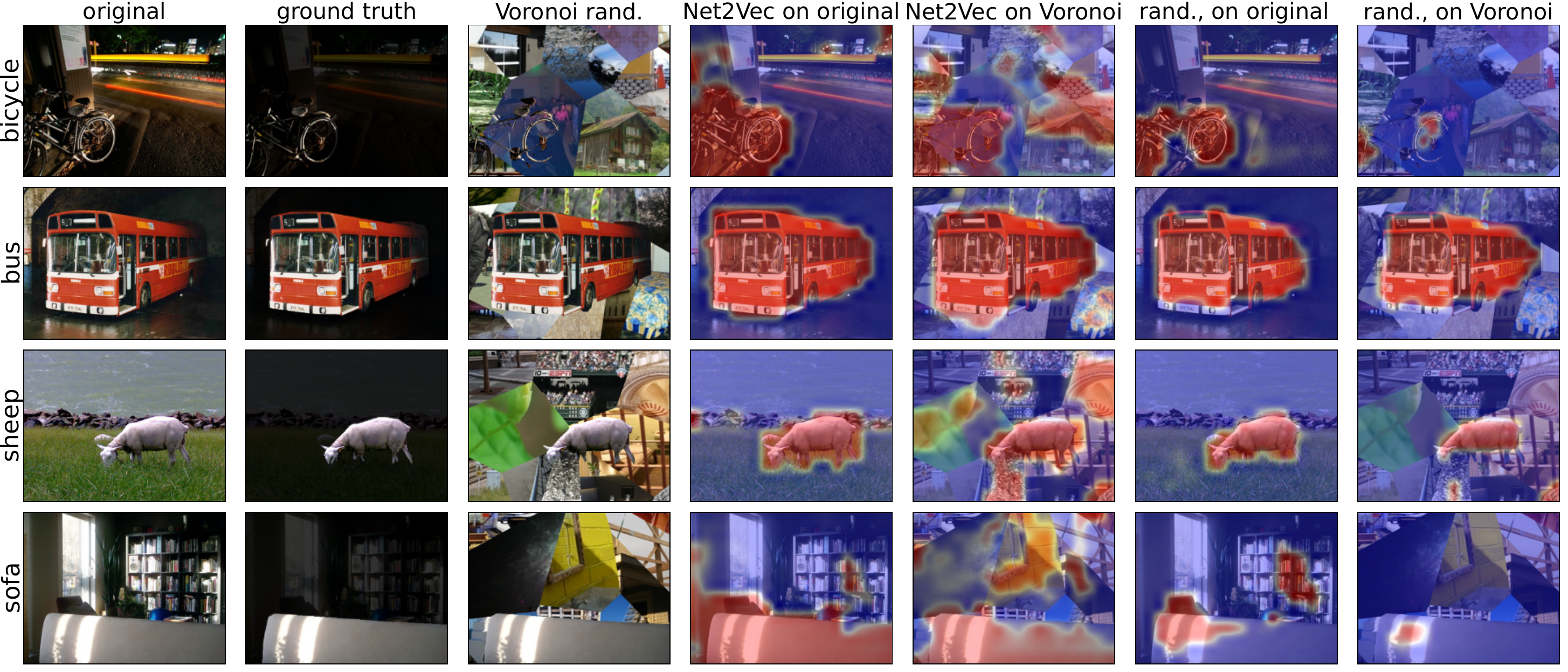}
    \end{subfigure}
    \begin{subfigure}{\linewidth}
        \caption{Exemplary inference results of \textbf{GloCE}s.}
        \includegraphics[width=\linewidth]{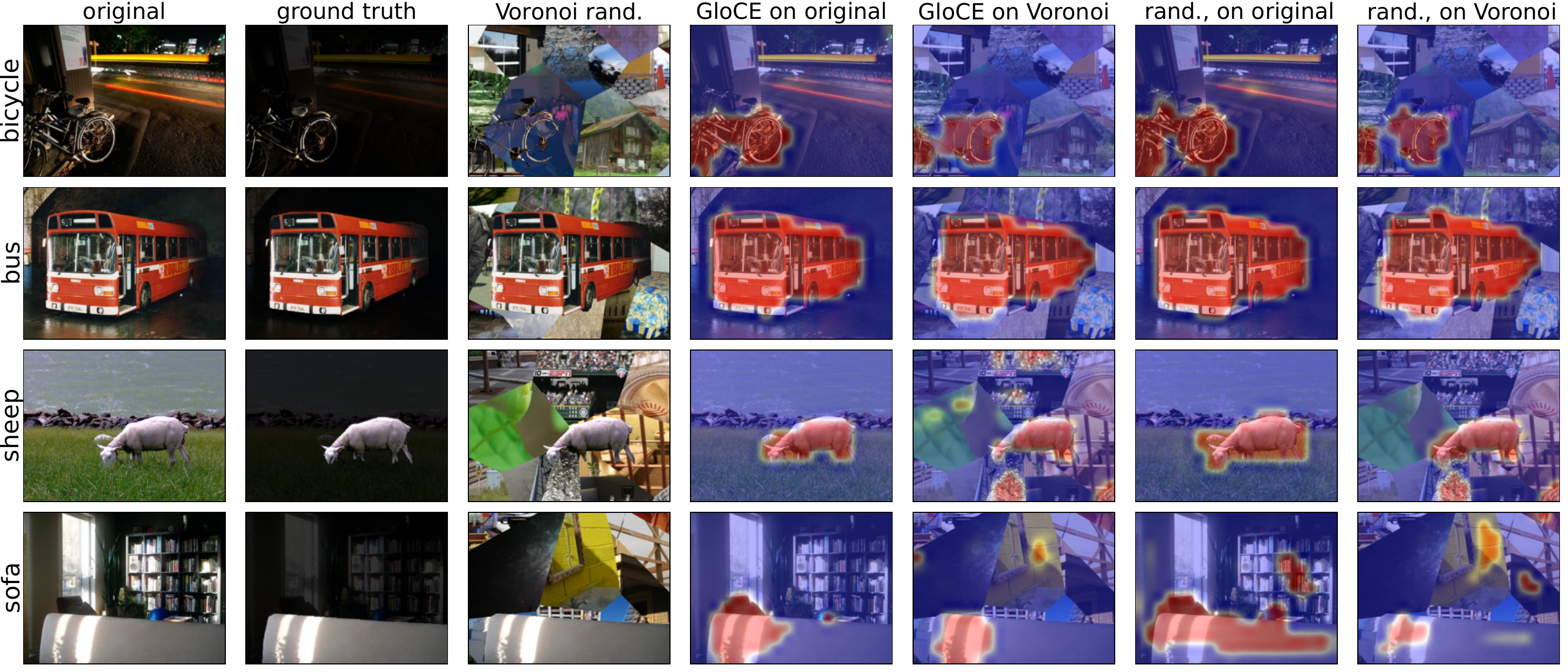}
    \end{subfigure}
    \caption{Exemplary results for Net2Vec CE and GloCEs and $4$ Pascal VOC concepts. For each concept, the heatmaps resulting from inference of a CE trained on vanilla data and one trained with simple Places205 background randomization (rand.) are shown side-by-side,
    each for the vanilla original and a Voronoi randomized version of a (randomly chosen) test sample.
    Ground truth masks (\textit{2nd column}) and predicted heatmaps (\textit{columns 4--7}) are shown via overlays: \textit{dark}/\textit{blue} means 0, \textit{no darkening}/\textit{red} means 1.
    \autoref{fig:voronoi} visualizes all three considered background randomization techniques.
    }
    \label{fig:examples}
\end{figure}

In this section, we detail the exact setup and results of our background robustness analysis.
The latter is split into three parts, following the questions posed in \autoref{sec:approach}:
(1) Whether a \emph{performance} drop is measurable when testing on manipulated background distributions \textit{(\emph{yes}, for some concepts and backgrounds)},
(2) whether training on background randomized samples results in different \emph{concept representations} \textit{(\emph{yes}, they do quite strongly)}, and lastly
(3) in how far the background-randomized training changes the performance of CEs \textit{(\emph{visibly} beneficial at low cost)}.
In the latter, we include an ablation study showing that neither a large number of randomized image variants is necessary, nor more than one layer per model in order to obtain relevant results.
Some examples of CE inference are provided in \autoref{fig:examples}.

\subsection{Experiment Settings}

\subsubsection{Concept Datasets.} As previously mentioned, we use the ImageNetS50 dataset \cite{gao2022luss}, which features segmentations of 50 object classes, each with 10 carefully selected training and 10 validation images. We note that CE training is few shot on images: The CEs have a very small amount of to-be-trained parameters (typically $\leq$ 500), while each single image gives rise to an activation map of spatial size at least $16\times 16 = 256$ pixels, each serving as one CE training input (cf.\ \autoref{fig:ce}). This amounts to at least $5\times$ the number of samples compared to the number of parameters in the vanilla case.
Apart from that, we also use a subset of 20 concepts from the Pascal VOC dataset with each 50 / 20 randomly selected train / validation samples (see \iouresultstables{} for the class list).
For evaluation on background-randomized test samples, each 
10 variants per foreground are created by random background sampling.

\subsubsection{DNNs and Layers.}
We follow prior work on LoCEs\footnote{\tiny\url{https://github.com/continental/local-concept-embeddings}} in \cite{mikriukov2024local}, regarding the choice of a diverse set of CNN and Vision Transformer (ViT) architectures, using: CNN classifiers VGG16\footnote{\label{footnote:torch-zoo}\tiny\url{https://pytorch.org/vision/stable/models}}~\cite{simonyan2015very}, MobileNetV3-L\textsuperscript{\ref{footnote:torch-zoo}}~\cite{howard2019searching} (MobileNet), and EfficientNet-B0\textsuperscript{\ref{footnote:torch-zoo}}~\cite{tan2019efficientnet} (EffNet); classification transformers ViT-B-16\textsuperscript{\ref{footnote:torch-zoo}}~\cite{dosovitskiy2021vit} (ViT), SWIN-T\textsuperscript{\ref{footnote:torch-zoo}}~\cite{liu2021swin} (SWIN); and object detectors YOLOv5s\footnote{\tiny\url{https://pytorch.org/hub/ultralytics_yolov5/}}~\cite{yolov5} (YOLO), with
residual backbone, and DETR\footnote{\tiny\url{https://huggingface.co/facebook/detr-resnet-50}}~\cite{carion2020detr} with ResNet50~\cite{simonyan2014very} backbone. For each we select an early, middle, and late layer to process (see \autoref{tab:layers}).

\subsubsection{CE Training.}
We consider two major paradigms for concept segmentation: The global one, derived from Net2Vec \cite{fong2018net2vec}, and the local one from the LoCE framework \cite{mikriukov2024local}. As a third set, we derive from the LoCEs their respective GloCE where necessary for evaluation.
Net2Vec CEs are trained with weighted binary cross-entropy loss (as compared to the less stable original intersection-based loss \cite{schwalbe2021verification}) with a weighting factor for class balancing as defined in the original paper~\cite{fong2018net2vec}, AdamW optimizer \cite{loshchilov2018decoupled} at batch size of
512 (LoCE) or 256 (Net2Vec) for 30 (ImageNet) or 20 (Pascal VOC) epochs.
As tested in \cite{schwalbe2021verification} and \cite{mikriukov2024local}, we skip the activation map thresholding from the original NetDissect approach \cite{bau2017network}, and reduce costs further by not upscaling the activation maps to full image size, but instead scaling both activations and ground truth masks to the common size of $80\times 80$ pixels (comparable to the choice made in \cite{mikriukov2024local}).
For the vanilla data, each CE is trained on 50 samples; and for the background-randomized samples, we explore generating each 1, 4, 8, and 32 variants per image (results in \autoref{fig:iou-num-bgs}).
LoCEs are trained with the same settings except for batch size, which is 1 since every LoCE only is trained on one input sample.

\begin{table}[tb]
    \centering
    \vspace*{-.5\baselineskip}
    \caption{Overview of used early, middle, and last layers per DNN.
    Used shorthands: m=model, bb=backbone, enc=encoder, f=features.}
    \scriptsize%
    \resizebox{.95\linewidth}{!}{%
    \begin{tabular*}{\linewidth}{@{}l >{\ttfamily}c @{\extracolsep{\fill}} >{\ttfamily}c @{\extracolsep{\fill}} >{\ttfamily}c@{}}\toprule
     & \textbf{\normalfont early}  & \textbf{\normalfont middle} & \textbf{\normalfont late}\\\midrule[\heavyrulewidth]
        detr:   &    m.bb.conv\_enc.m.layer4 & m.input\_projection & m.enc.layers.5 \\\midrule
        vit:   &    conv\_proj & enc.layers.encoder\_layer\_6 & enc.layers.encoder\_layer\_11 \\\midrule
        swin:   &    f.0 & f.3 & features.7 \\\midrule
        efficientnet:   &    f.4.2 & f.6.0 & f.7.0 \\\midrule
        mobilenet:   &    f.7.block.2.0 & f.12.block.3.0 & f.16.0 \\\midrule
        yolo:   &    4.cv3.conv & 14.conv & 23.cv3.conv \\\midrule
        vgg:   &    f.7 & f.21 & f.28 \\\bottomrule
    \end{tabular*}
    }
    \vspace*{-\baselineskip}
    \label{tab:layers}
\end{table}

\subsection{Testing for Background Bias}\label{sec:exp-testing}

One of our main questions is whether the (foreground) performance of a CE depends on the backgrounds.
To answer this, we test the generalization performance of the CEs on images with foregrounds from the concept test set, but backgrounds randomized using images from the Places dataset.
This allows to compare IoU performance for:
\begin{itemize}
\item the unchanged original image background (\textsf{vanilla});
\item each of the defined Places background superclasses (see \autoref{tab:places-supclasses}); and
\item on an approximately uniform sample of backgrounds, serving as the baseline (denoted as \textsf{any} category).
\end{itemize}
If a CE's performance on an individual background class decreases compared to \textsf{any} background this means that the CE is negatively biased with respect to that background (detecting it there is harder); and vice versa.
This captures whether a background faces suspiciously bad (or good) performance compared to the other backgrounds.
\iouresultstables{} show for each pair of concept and background class the relative change of IoU results compared to the \textsf{any} results for that concept (IoU values averaged over all test samples and late layer CEs of all models).
Note that the vanilla test values could also serve as a baseline here. This would, however, not change the \textit{relative} differences.

\begin{table}[tbp]
\scriptsize
\sisetup{table-format=+2.1}

\caption{\textbf{Background bias (Pascal VOC concepts):} The values show in \% how much the average test IoU of vanilla CEs increases (marked \emph{red}) respectively decreases (marked \emph{blue}) on specific background categories compared to performance on arbitrary backgrounds. Test samples are created via simple background pasting ($4$ random variants per test sample). CEs are global Net2Vec CEs (\emph{top}) respectively globalized LoCEs (\emph{bottom}) for the shown Pascal VOC concepts and late model layers, IoUs averaged over $7$ models.
}\label{tab:iou-per-bg-cat-voc}
\renewcommand\arraystretch{1.2}
\newcommand{\tabh}[1]{\multicolumn{1}{c}{\parbox[t]{3.2em}{\centering\sffamily #1}}}
\resizebox{\linewidth}{!}{%
\begin{subtable}{\linewidth}
\caption{\textbf{Net2Vec} results}
\label{tab:iou-per-bg-cat-voc-net2vec}
\centering%
\color{black}%
\begin{tabular}{@{}>{\sffamily}lSSSSSSSSSSS@{}}
VOC Concept & \tabh{arch.} & \tabh{at water} & \tabh{botanical} & \tabh{field} & \tabh{forest} & \tabh{indoors} & \tabh{mach.} & \tabh{open l.} & \tabh{road} & \tabh{snow} & \tabh{vanilla} \\\midrule[\heavyrulewidth]
aeroplane & \cellcolor[HTML]{F9EEE7} -3.43 & \cellcolor[HTML]{D7E8F1} 8.45 & \cellcolor[HTML]{F8F3F0} -1.53 & \cellcolor[HTML]{CAE1EE} 11.26 & \cellcolor[HTML]{F8F4F2} -1.01 & \cellcolor[HTML]{F3A481} -19.92 & \cellcolor[HTML]{DF765E}\color[HTML]{F1F1F1}-26.90 & \cellcolor[HTML]{C5DFEC} 12.07 & \cellcolor[HTML]{F7F6F6} -0.19 & \cellcolor[HTML]{C7E0ED} 11.65 & \cellcolor[HTML]{BDDBEA} 12.98 \\
bicycle & \cellcolor[HTML]{F9EEE7} -3.13 & \cellcolor[HTML]{C7E0ED} 11.62 & \cellcolor[HTML]{F9F0EB} -2.38 & \cellcolor[HTML]{CCE2EF} 10.92 & \cellcolor[HTML]{F3F5F6} 0.95 & \cellcolor[HTML]{FDDBC7} -10.13 & \cellcolor[HTML]{C43B3C}\color[HTML]{F1F1F1}-35.00 & \cellcolor[HTML]{9DCBE1} 18.19 & \cellcolor[HTML]{F6F7F7} 0.29 & \cellcolor[HTML]{D2E6F0} 9.46 & \cellcolor[HTML]{FBCEB7} -12.33 \\
bird & \cellcolor[HTML]{F8F4F2} -0.84 & \cellcolor[HTML]{E3EDF3} 5.40 & \cellcolor[HTML]{F0F4F6} 1.84 & \cellcolor[HTML]{DBEAF2} 7.07 & \cellcolor[HTML]{CCE2EF} 10.75 & \cellcolor[HTML]{FCDECD} -8.95 & \cellcolor[HTML]{F5AC8B} -18.70 & \cellcolor[HTML]{DBEAF2} 7.37 & \cellcolor[HTML]{F8F1ED} -1.97 & \cellcolor[HTML]{E9F0F4} 3.79 & \cellcolor[HTML]{87BEDA} 21.19 \\
boat & \cellcolor[HTML]{FCD5BF} -11.32 & \cellcolor[HTML]{B8D8E9} 13.98 & \cellcolor[HTML]{E7F0F4} 4.22 & \cellcolor[HTML]{9DCBE1} 18.03 & \cellcolor[HTML]{B8D8E9} 14.01 & \cellcolor[HTML]{E98B6E}\color[HTML]{F1F1F1}-23.55 & \cellcolor[HTML]{BA2832}\color[HTML]{F1F1F1}-37.89 & \cellcolor[HTML]{87BEDA} 21.12 & \cellcolor[HTML]{FCD7C2} -10.92 & \cellcolor[HTML]{E7F0F4} 3.91 & \cellcolor[HTML]{C7E0ED} 11.46 \\
bottle & \cellcolor[HTML]{F9EFE9} -2.75 & \cellcolor[HTML]{DEEBF2} 6.61 & \cellcolor[HTML]{EFF3F5} 2.24 & \cellcolor[HTML]{D5E7F1} 8.80 & \cellcolor[HTML]{D7E8F1} 8.45 & \cellcolor[HTML]{FDDCC9} -9.55 & \cellcolor[HTML]{D7634F}\color[HTML]{F1F1F1}-29.56 & \cellcolor[HTML]{ACD2E5} 15.83 & \cellcolor[HTML]{FCE0D0} -8.15 & \cellcolor[HTML]{E0ECF3} 6.21 & \cellcolor[HTML]{F6AF8E} -18.10 \\
bus & \cellcolor[HTML]{FAEAE1} -4.56 & \cellcolor[HTML]{F0F4F6} 1.89 & \cellcolor[HTML]{EFF3F5} 2.03 & \cellcolor[HTML]{E4EEF4} 4.75 & \cellcolor[HTML]{E9F0F4} 3.76 & \cellcolor[HTML]{F9C2A7} -14.84 & \cellcolor[HTML]{EF9979} -21.54 & \cellcolor[HTML]{E9F0F4} 3.75 & \cellcolor[HTML]{F9EEE7} -3.42 & \cellcolor[HTML]{EDF2F5} 2.50 & \cellcolor[HTML]{F6F7F7} 0.03 \\
car & \cellcolor[HTML]{FAE7DC} -5.75 & \cellcolor[HTML]{E3EDF3} 5.42 & \cellcolor[HTML]{F0F4F6} 1.79 & \cellcolor[HTML]{CCE2EF} 10.85 & \cellcolor[HTML]{D5E7F1} 8.63 & \cellcolor[HTML]{F6B191} -17.81 & \cellcolor[HTML]{DD7059}\color[HTML]{F1F1F1}-27.54 & \cellcolor[HTML]{CFE4EF} 10.32 & \cellcolor[HTML]{FBE4D6} -6.81 & \cellcolor[HTML]{E6EFF4} 4.63 & \cellcolor[HTML]{FBE6DA} -6.22 \\
cat & \cellcolor[HTML]{EFF3F5} 2.24 & \cellcolor[HTML]{F8F4F2} -1.06 & \cellcolor[HTML]{F9F0EB} -2.48 & \cellcolor[HTML]{F8F2EF} -1.89 & \cellcolor[HTML]{F8F2EF} -1.73 & \cellcolor[HTML]{D7E8F1} 8.41 & \cellcolor[HTML]{F7F6F6} -0.32 & \cellcolor[HTML]{F9EBE3} -4.15 & \cellcolor[HTML]{EFF3F5} 2.08 & \cellcolor[HTML]{F8F1ED} -2.14 & \cellcolor[HTML]{CCE2EF} 10.61 \\
chair & \cellcolor[HTML]{FAC8AF} -13.53 & \cellcolor[HTML]{CCE2EF} 10.85 & \cellcolor[HTML]{EAF1F5} 3.43 & \cellcolor[HTML]{B8D8E9} 13.89 & \cellcolor[HTML]{ACD2E5} 15.98 & \cellcolor[HTML]{E6866A}\color[HTML]{F1F1F1}-24.60 & \cellcolor[HTML]{C6413E}\color[HTML]{F1F1F1}-34.30 & \cellcolor[HTML]{CAE1EE} 11.30 & \cellcolor[HTML]{FBE5D8} -6.56 & \cellcolor[HTML]{CFE4EF} 10.42 & \cellcolor[HTML]{F7B596} -16.95 \\
cow & \cellcolor[HTML]{F9EDE5} -3.64 & \cellcolor[HTML]{E0ECF3} 6.05 & \cellcolor[HTML]{F2F5F6} 1.30 & \cellcolor[HTML]{D5E7F1} 8.78 & \cellcolor[HTML]{E4EEF4} 4.86 & \cellcolor[HTML]{F5AA89} -19.05 & \cellcolor[HTML]{F6B191} -17.87 & \cellcolor[HTML]{E3EDF3} 5.44 & \cellcolor[HTML]{F6F7F7} 0.19 & \cellcolor[HTML]{D8E9F1} 7.86 & \cellcolor[HTML]{B3D6E8} 14.66 \\
dining table & \cellcolor[HTML]{F9EFE9} -2.76 & \cellcolor[HTML]{F8F4F2} -0.98 & \cellcolor[HTML]{F8F4F2} -1.17 & \cellcolor[HTML]{F7F6F6} -0.17 & \cellcolor[HTML]{EFF3F5} 2.13 & \cellcolor[HTML]{FAE8DE} -5.30 & \cellcolor[HTML]{F9F0EB} -2.69 & \cellcolor[HTML]{F8F3F0} -1.29 & \cellcolor[HTML]{F9EEE7} -3.42 & \cellcolor[HTML]{F8F4F2} -0.97 & \cellcolor[HTML]{EDF2F5} 2.59 \\
dog & \cellcolor[HTML]{F8F1ED} -2.30 & \cellcolor[HTML]{EAF1F5} 3.13 & \cellcolor[HTML]{EDF2F5} 2.55 & \cellcolor[HTML]{EAF1F5} 3.45 & \cellcolor[HTML]{E9F0F4} 3.64 & \cellcolor[HTML]{FCE2D2} -7.56 & \cellcolor[HTML]{FACAB1} -13.17 & \cellcolor[HTML]{ECF2F5} 2.75 & \cellcolor[HTML]{F8F4F2} -1.05 & \cellcolor[HTML]{E7F0F4} 4.29 & \cellcolor[HTML]{F9F0EB} -2.55 \\
horse & \cellcolor[HTML]{F8F3F0} -1.20 & \cellcolor[HTML]{ECF2F5} 2.94 & \cellcolor[HTML]{F0F4F6} 1.92 & \cellcolor[HTML]{D8E9F1} 7.95 & \cellcolor[HTML]{E1EDF3} 5.82 & \cellcolor[HTML]{FBCEB7} -12.46 & \cellcolor[HTML]{F8BFA4} -15.12 & \cellcolor[HTML]{E7F0F4} 4.00 & \cellcolor[HTML]{F7F6F6} -0.05 & \cellcolor[HTML]{F5F6F7} 0.46 & \cellcolor[HTML]{D8E9F1} 7.94 \\
motorbike & \cellcolor[HTML]{FBE5D8} -6.31 & \cellcolor[HTML]{E1EDF3} 5.68 & \cellcolor[HTML]{E4EEF4} 4.84 & \cellcolor[HTML]{D4E6F1} 9.37 & \cellcolor[HTML]{E9F0F4} 3.67 & \cellcolor[HTML]{FCD3BC} -11.66 & \cellcolor[HTML]{F7B799} -16.75 & \cellcolor[HTML]{D4E6F1} 9.02 & \cellcolor[HTML]{F7F6F6} -0.20 & \cellcolor[HTML]{E7F0F4} 4.24 & \cellcolor[HTML]{A9D1E5} 16.17 \\
person & \cellcolor[HTML]{F5F6F7} 0.46 & \cellcolor[HTML]{F7F6F6} -0.20 & \cellcolor[HTML]{ECF2F5} 2.78 & \cellcolor[HTML]{F3F5F6} 1.06 & \cellcolor[HTML]{F2F5F6} 1.34 & \cellcolor[HTML]{F8F2EF} -1.68 & \cellcolor[HTML]{FBE4D6} -6.85 & \cellcolor[HTML]{F8F1ED} -2.33 & \cellcolor[HTML]{F8F1ED} -2.25 & \cellcolor[HTML]{F8F1ED} -2.05 & \cellcolor[HTML]{FBE6DA} -6.25 \\
potted plant & \cellcolor[HTML]{F0F4F6} 1.68 & \cellcolor[HTML]{EFF3F5} 2.23 & \cellcolor[HTML]{F7B799} -16.46 & \cellcolor[HTML]{F3F5F6} 0.79 & \cellcolor[HTML]{FBE5D8} -6.50 & \cellcolor[HTML]{EFF3F5} 2.04 & \cellcolor[HTML]{FCD7C2} -10.67 & \cellcolor[HTML]{D4E6F1} 9.16 & \cellcolor[HTML]{EDF2F5} 2.43 & \cellcolor[HTML]{DBEAF2} 7.37 & \cellcolor[HTML]{F7F6F6} -0.37 \\
sheep & \cellcolor[HTML]{F8F1ED} -2.30 & \cellcolor[HTML]{ECF2F5} 3.08 & \cellcolor[HTML]{F3F5F6} 1.06 & \cellcolor[HTML]{DDEBF2} 6.96 & \cellcolor[HTML]{E1EDF3} 5.61 & \cellcolor[HTML]{FBE3D4} -7.35 & \cellcolor[HTML]{FDD9C4} -10.42 & \cellcolor[HTML]{F7F5F4} -0.61 & \cellcolor[HTML]{F2F5F6} 1.20 & \cellcolor[HTML]{F8F4F2} -1.16 & \cellcolor[HTML]{A2CDE3} 17.52 \\
sofa & \cellcolor[HTML]{F7F6F6} -0.31 & \cellcolor[HTML]{F8F4F2} -0.91 & \cellcolor[HTML]{EFF3F5} 2.17 & \cellcolor[HTML]{F0F4F6} 1.59 & \cellcolor[HTML]{E9F0F4} 3.72 & \cellcolor[HTML]{EFF3F5} 2.22 & \cellcolor[HTML]{FCDFCF} -8.39 & \cellcolor[HTML]{FAE8DE} -5.43 & \cellcolor[HTML]{F8F1ED} -2.21 & \cellcolor[HTML]{FCE2D2} -7.59 & \cellcolor[HTML]{CAE1EE} 10.98 \\
train & \cellcolor[HTML]{FCE2D2} -7.81 & \cellcolor[HTML]{DEEBF2} 6.44 & \cellcolor[HTML]{F5F6F7} 0.55 & \cellcolor[HTML]{B3D6E8} 14.82 & \cellcolor[HTML]{E9F0F4} 3.65 & \cellcolor[HTML]{EB9172}\color[HTML]{F1F1F1}-22.76 & \cellcolor[HTML]{CC4C44}\color[HTML]{F1F1F1}-32.48 & \cellcolor[HTML]{A7D0E4} 16.62 & \cellcolor[HTML]{FAE9DF} -4.91 & \cellcolor[HTML]{C7E0ED} 11.40 & \cellcolor[HTML]{C2DDEC} 12.15 \\
TV monitor & \cellcolor[HTML]{F2F5F6} 1.40 & \cellcolor[HTML]{F8F1ED} -2.18 & \cellcolor[HTML]{F8F4F2} -1.12 & \cellcolor[HTML]{ECF2F5} 3.07 & \cellcolor[HTML]{EFF3F5} 2.12 & \cellcolor[HTML]{E3EDF3} 5.42 & \cellcolor[HTML]{F6F7F7} 0.03 & \cellcolor[HTML]{F0F4F6} 1.61 & \cellcolor[HTML]{F8F3F0} -1.19 & \cellcolor[HTML]{F6F7F7} 0.18 & \cellcolor[HTML]{E6EFF4} 4.52 \\
\bottomrule
\end{tabular}
\end{subtable}
}%
\\\resizebox{\linewidth}{!}{%
\begin{subtable}{\linewidth}
\caption{\textbf{GloCE} results}
\label{tab:iou-per-bg-cat-voc-gloce}
\centering%
\color{black}%
\begin{tabular}{>{\sffamily}lSSSSSSSSSSS}
VOC Concept & \tabh{arch.} & \tabh{at water} & \tabh{botanical} & \tabh{field} & \tabh{forest} & \tabh{indoors} & \tabh{mach.} & \tabh{open l.} & \tabh{road} & \tabh{snow} & \tabh{vanilla} \\\midrule[\heavyrulewidth]
aeroplane & \cellcolor[HTML]{FAE8DE} -5.22 & \cellcolor[HTML]{DBEAF2} 7.33 & \cellcolor[HTML]{ECF2F5} 2.86 & \cellcolor[HTML]{C5DFEC} 12.08 & \cellcolor[HTML]{D2E6F0} 9.45 & \cellcolor[HTML]{F8BFA4} -14.94 & \cellcolor[HTML]{CF5246}\color[HTML]{F1F1F1}-31.94 & \cellcolor[HTML]{CCE2EF} 10.75 & \cellcolor[HTML]{F8F2EF} -1.74 & \cellcolor[HTML]{E1EDF3} 5.54 & \cellcolor[HTML]{A2CDE3} 17.30 \\
bicycle & \cellcolor[HTML]{FCE2D2} -7.45 & \cellcolor[HTML]{E0ECF3} 6.13 & \cellcolor[HTML]{ECF2F5} 3.02 & \cellcolor[HTML]{C2DDEC} 12.49 & \cellcolor[HTML]{E3EDF3} 5.40 & \cellcolor[HTML]{F9EBE3} -4.26 & \cellcolor[HTML]{E58368}\color[HTML]{F1F1F1}-24.77 & \cellcolor[HTML]{CAE1EE} 11.02 & \cellcolor[HTML]{E7F0F4} 4.25 & \cellcolor[HTML]{F7F5F4} -0.76 & \cellcolor[HTML]{AED3E6} 15.53 \\
bird & \cellcolor[HTML]{FAE9DF} -4.69 & \cellcolor[HTML]{D2E6F0} 9.41 & \cellcolor[HTML]{D5E7F1} 8.79 & \cellcolor[HTML]{C0DCEB} 12.62 & \cellcolor[HTML]{84BCD9} 21.83 & \cellcolor[HTML]{F9C6AC} -13.78 & \cellcolor[HTML]{F5AC8B} -18.61 & \cellcolor[HTML]{ACD2E5} 15.76 & \cellcolor[HTML]{F8F1ED} -2.17 & \cellcolor[HTML]{E9F0F4} 3.70 & \cellcolor[HTML]{4291C2}\color[HTML]{F1F1F1}30.17 \\
boat & \cellcolor[HTML]{FBD0B9} -11.76 & \cellcolor[HTML]{9BC9E0} 18.54 & \cellcolor[HTML]{E0ECF3} 5.91 & \cellcolor[HTML]{8DC2DC} 20.33 & \cellcolor[HTML]{AED3E6} 15.61 & \cellcolor[HTML]{DD7059}\color[HTML]{F1F1F1}-27.56 & \cellcolor[HTML]{991027}\color[HTML]{F1F1F1}-43.19 & \cellcolor[HTML]{8DC2DC} 20.59 & \cellcolor[HTML]{F9EDE5} -3.83 & \cellcolor[HTML]{DEEBF2} 6.59 & \cellcolor[HTML]{81BAD8} 22.18 \\
bottle & \cellcolor[HTML]{FBCEB7} -12.42 & \cellcolor[HTML]{BBDAEA} 13.62 & \cellcolor[HTML]{C7E0ED} 11.44 & \cellcolor[HTML]{A2CDE3} 17.25 & \cellcolor[HTML]{9BC9E0} 18.52 & \cellcolor[HTML]{F9EDE5} -3.89 & \cellcolor[HTML]{D7634F}\color[HTML]{F1F1F1}-29.34 & \cellcolor[HTML]{529DC8}\color[HTML]{F1F1F1}27.99 & \cellcolor[HTML]{F9F0EB} -2.64 & \cellcolor[HTML]{B6D7E8} 14.34 & \cellcolor[HTML]{75B2D4}\color[HTML]{F1F1F1}23.80 \\
bus & \cellcolor[HTML]{FAE7DC} -5.51 & \cellcolor[HTML]{EDF2F5} 2.57 & \cellcolor[HTML]{EFF3F5} 2.14 & \cellcolor[HTML]{E0ECF3} 6.20 & \cellcolor[HTML]{E9F0F4} 3.70 & \cellcolor[HTML]{FCD7C2} -10.90 & \cellcolor[HTML]{F7B596} -17.12 & \cellcolor[HTML]{E9F0F4} 3.84 & \cellcolor[HTML]{F7F5F4} -0.66 & \cellcolor[HTML]{EFF3F5} 2.28 & \cellcolor[HTML]{E7F0F4} 4.13 \\
car & \cellcolor[HTML]{FCE0D0} -8.05 & \cellcolor[HTML]{E6EFF4} 4.65 & \cellcolor[HTML]{E3EDF3} 5.45 & \cellcolor[HTML]{C0DCEB} 12.84 & \cellcolor[HTML]{CCE2EF} 10.75 & \cellcolor[HTML]{F8BDA1} -15.59 & \cellcolor[HTML]{EC9374}\color[HTML]{F1F1F1}-22.29 & \cellcolor[HTML]{CCE2EF} 10.68 & \cellcolor[HTML]{F9F0EB} -2.45 & \cellcolor[HTML]{E9F0F4} 3.64 & \cellcolor[HTML]{B3D6E8} 14.53 \\
cat & \cellcolor[HTML]{F7F5F4} -0.76 & \cellcolor[HTML]{F8F4F2} -0.85 & \cellcolor[HTML]{F9EDE5} -3.60 & \cellcolor[HTML]{F8F3F0} -1.53 & \cellcolor[HTML]{F7F5F4} -0.53 & \cellcolor[HTML]{DDEBF2} 6.86 & \cellcolor[HTML]{F7F5F4} -0.50 & \cellcolor[HTML]{FBE6DA} -6.23 & \cellcolor[HTML]{F5F6F7} 0.53 & \cellcolor[HTML]{F8F4F2} -1.09 & \cellcolor[HTML]{3E8CBF}\color[HTML]{F1F1F1}31.45 \\
chair & \cellcolor[HTML]{FCD5BF} -11.04 & \cellcolor[HTML]{C2DDEC} 12.13 & \cellcolor[HTML]{A9D1E5} 16.13 & \cellcolor[HTML]{90C4DD} 20.07 & \cellcolor[HTML]{7EB8D7} 22.32 & \cellcolor[HTML]{E98B6E}\color[HTML]{F1F1F1}-23.69 & \cellcolor[HTML]{B61F2E}\color[HTML]{F1F1F1}-38.75 & \cellcolor[HTML]{ACD2E5} 15.84 & \cellcolor[HTML]{DDEBF2} 6.70 & \cellcolor[HTML]{E1EDF3} 5.48 & \cellcolor[HTML]{840924}\color[HTML]{F1F1F1}-45.87 \\
cow & \cellcolor[HTML]{FAE9DF} -4.91 & \cellcolor[HTML]{EDF2F5} 2.41 & \cellcolor[HTML]{F7F6F6} -0.24 & \cellcolor[HTML]{DAE9F2} 7.56 & \cellcolor[HTML]{E3EDF3} 5.36 & \cellcolor[HTML]{F9C4A9} -14.44 & \cellcolor[HTML]{EB9172}\color[HTML]{F1F1F1}-22.92 & \cellcolor[HTML]{E6EFF4} 4.45 & \cellcolor[HTML]{F5F6F7} 0.50 & \cellcolor[HTML]{F2F5F6} 1.54 & \cellcolor[HTML]{A0CCE2} 17.71 \\
dining table & \cellcolor[HTML]{F9EBE3} -4.22 & \cellcolor[HTML]{F8F4F2} -0.84 & \cellcolor[HTML]{EFF3F5} 2.24 & \cellcolor[HTML]{EAF1F5} 3.23 & \cellcolor[HTML]{D8E9F1} 8.12 & \cellcolor[HTML]{F6F7F7} 0.35 & \cellcolor[HTML]{FBE4D6} -6.70 & \cellcolor[HTML]{F8F3F0} -1.42 & \cellcolor[HTML]{FBE6DA} -6.21 & \cellcolor[HTML]{FBE5D8} -6.46 & \cellcolor[HTML]{A9D1E5} 16.10 \\
dog & \cellcolor[HTML]{F9F0EB} -2.69 & \cellcolor[HTML]{EDF2F5} 2.68 & \cellcolor[HTML]{F2F5F6} 1.33 & \cellcolor[HTML]{E4EEF4} 4.86 & \cellcolor[HTML]{E7F0F4} 4.16 & \cellcolor[HTML]{F9EFE9} -2.98 & \cellcolor[HTML]{FAC8AF} -13.41 & \cellcolor[HTML]{F0F4F6} 1.93 & \cellcolor[HTML]{F6F7F7} 0.22 & \cellcolor[HTML]{EDF2F5} 2.57 & \cellcolor[HTML]{DBEAF2} 7.41 \\
horse & \cellcolor[HTML]{F7F6F6} -0.16 & \cellcolor[HTML]{EFF3F5} 2.16 & \cellcolor[HTML]{EFF3F5} 2.02 & \cellcolor[HTML]{DBEAF2} 7.04 & \cellcolor[HTML]{E6EFF4} 4.60 & \cellcolor[HTML]{FDDBC7} -10.07 & \cellcolor[HTML]{F6B394} -17.42 & \cellcolor[HTML]{E9F0F4} 3.66 & \cellcolor[HTML]{F0F4F6} 1.87 & \cellcolor[HTML]{F8F4F2} -0.96 & \cellcolor[HTML]{D2E6F0} 9.54 \\
motorbike & \cellcolor[HTML]{F8F4F2} -1.06 & \cellcolor[HTML]{E0ECF3} 6.22 & \cellcolor[HTML]{ECF2F5} 3.06 & \cellcolor[HTML]{D4E6F1} 9.29 & \cellcolor[HTML]{E3EDF3} 5.40 & \cellcolor[HTML]{FBE3D4} -7.07 & \cellcolor[HTML]{F09C7B} -21.33 & \cellcolor[HTML]{CFE4EF} 10.36 & \cellcolor[HTML]{F5F6F7} 0.40 & \cellcolor[HTML]{E7F0F4} 4.13 & \cellcolor[HTML]{E3EDF3} 5.25 \\
person & \cellcolor[HTML]{ECF2F5} 2.75 & \cellcolor[HTML]{E6EFF4} 4.60 & \cellcolor[HTML]{E9F0F4} 3.78 & \cellcolor[HTML]{E9F0F4} 3.88 & \cellcolor[HTML]{EDF2F5} 2.51 & \cellcolor[HTML]{F9EBE3} -4.27 & \cellcolor[HTML]{FCD7C2} -10.70 & \cellcolor[HTML]{EFF3F5} 1.96 & \cellcolor[HTML]{F0F4F6} 1.90 & \cellcolor[HTML]{F7F6F6} -0.30 & \cellcolor[HTML]{F9C2A7} -14.76 \\
potted plant & \cellcolor[HTML]{F8F3F0} -1.31 & \cellcolor[HTML]{F2F5F6} 1.26 & \cellcolor[HTML]{F19E7D} -20.94 & \cellcolor[HTML]{F8F2EF} -1.62 & \cellcolor[HTML]{FAC8AF} -13.39 & \cellcolor[HTML]{DBEAF2} 7.07 & \cellcolor[HTML]{F9EBE3} -3.92 & \cellcolor[HTML]{DBEAF2} 7.41 & \cellcolor[HTML]{DBEAF2} 7.23 & \cellcolor[HTML]{E0ECF3} 6.00 & \cellcolor[HTML]{FDDCC9} -9.68 \\
sheep & \cellcolor[HTML]{FDDCC9} -9.62 & \cellcolor[HTML]{E7F0F4} 4.08 & \cellcolor[HTML]{F5F6F7} 0.66 & \cellcolor[HTML]{CAE1EE} 11.15 & \cellcolor[HTML]{DDEBF2} 6.97 & \cellcolor[HTML]{F8BFA4} -15.01 & \cellcolor[HTML]{F3A481} -20.24 & \cellcolor[HTML]{E9F0F4} 3.81 & \cellcolor[HTML]{F8F1ED} -2.21 & \cellcolor[HTML]{ECF2F5} 2.79 & \cellcolor[HTML]{266CAF}\color[HTML]{F1F1F1}38.64 \\
sofa & \cellcolor[HTML]{FDDDCB} -9.18 & \cellcolor[HTML]{F6F7F7} 0.10 & \cellcolor[HTML]{E6EFF4} 4.34 & \cellcolor[HTML]{E0ECF3} 5.96 & \cellcolor[HTML]{D7E8F1} 8.54 & \cellcolor[HTML]{F7F6F6} -0.10 & \cellcolor[HTML]{E8896C}\color[HTML]{F1F1F1}-23.92 & \cellcolor[HTML]{FBE3D4} -7.21 & \cellcolor[HTML]{F9F0EB} -2.37 & \cellcolor[HTML]{FCD7C2} -10.61 & \cellcolor[HTML]{307AB6}\color[HTML]{F1F1F1}35.46 \\
train & \cellcolor[HTML]{FBE5D8} -6.38 & \cellcolor[HTML]{EDF2F5} 2.41 & \cellcolor[HTML]{EDF2F5} 2.68 & \cellcolor[HTML]{D1E5F0} 9.97 & \cellcolor[HTML]{D7E8F1} 8.23 & \cellcolor[HTML]{F7B596} -16.90 & \cellcolor[HTML]{D35A4A}\color[HTML]{F1F1F1}-30.78 & \cellcolor[HTML]{CFE4EF} 10.53 & \cellcolor[HTML]{FAEAE1} -4.39 & \cellcolor[HTML]{EAF1F5} 3.44 & \cellcolor[HTML]{93C6DE} 19.65 \\
TV monitor & \cellcolor[HTML]{F9EDE5} -3.87 & \cellcolor[HTML]{F9EDE5} -3.57 & \cellcolor[HTML]{E3EDF3} 5.36 & \cellcolor[HTML]{EAF1F5} 3.43 & \cellcolor[HTML]{DBEAF2} 7.06 & \cellcolor[HTML]{E1EDF3} 5.74 & \cellcolor[HTML]{FCD5BF} -11.32 & \cellcolor[HTML]{EFF3F5} 2.24 & \cellcolor[HTML]{FAE8DE} -5.44 & \cellcolor[HTML]{F9EEE7} -3.49 & \cellcolor[HTML]{AED3E6} 15.60 \\
\bottomrule
\end{tabular}
\end{subtable}
}%
\end{table}

\begin{table}[tbp]
\scriptsize
\sisetup{table-format=+2.1}

\caption{\textbf{Background bias (ImageNetS50 concepts, Net2Vec):} The values show in \% how much the average test IoU of vanilla CEs increases (marked \emph{red}) respectively decreases (marked \emph{blue}) on specific background categories compared to performance on arbitrary backgrounds. Test samples are created via simple background pasting (4 random variants per test sample). CEs are global Net2Vec CEs for the shown ImageNetS concepts and late model layers, IoUs averaged over 7 models.
}\label{tab:iou-per-bg-cat-imagenet-net2vec}
\renewcommand\arraystretch{1.2}
\newcommand{\tabh}[1]{\multicolumn{1}{c}{\parbox[t]{3.2em}{\centering\sffamily #1}}}
\scriptsize
\resizebox{.98\linewidth}{!}{%
  \color{black}%
\begin{tabular}{@{}>{\sffamily}lSSSSSSSSSSS@{}}
VOC Concept & \tabh{arch.} & \tabh{at water} & \tabh{botanical} & \tabh{field} & \tabh{forest} & \tabh{indoors} & \tabh{mach.} & \tabh{open l.} & \tabh{road} & \tabh{snow} & \tabh{vanilla} \\\midrule[\heavyrulewidth]
African elephant & \cellcolor[HTML]{FBE6DA} -6.15 & \cellcolor[HTML]{F2F5F6} 1.36 & \cellcolor[HTML]{EFF3F5} 2.25 & \cellcolor[HTML]{E1EDF3} 5.50 & \cellcolor[HTML]{EAF1F5} 3.49 & \cellcolor[HTML]{FBCCB4} -12.59 & \cellcolor[HTML]{F8BB9E} -15.77 & \cellcolor[HTML]{F0F4F6} 1.83 & \cellcolor[HTML]{F8F1ED} -2.15 & \cellcolor[HTML]{F6F7F7} 0.07 & \cellcolor[HTML]{CFE4EF} 10.38 \\
agaric & \cellcolor[HTML]{F8F1ED} -2.30 & \cellcolor[HTML]{EDF2F5} 2.62 & \cellcolor[HTML]{F2F5F6} 1.44 & \cellcolor[HTML]{F0F4F6} 1.81 & \cellcolor[HTML]{DBEAF2} 7.29 & \cellcolor[HTML]{FBD0B9} -12.00 & \cellcolor[HTML]{FDDDCB} -9.21 & \cellcolor[HTML]{EDF2F5} 2.55 & \cellcolor[HTML]{F9EEE7} -3.31 & \cellcolor[HTML]{F6F7F7} 0.26 & \cellcolor[HTML]{C5DFEC} 11.83 \\
airliner & \cellcolor[HTML]{F7F6F6} -0.00 & \cellcolor[HTML]{E1EDF3} 5.83 & \cellcolor[HTML]{E1EDF3} 5.75 & \cellcolor[HTML]{D7E8F1} 8.25 & \cellcolor[HTML]{D5E7F1} 8.72 & \cellcolor[HTML]{FBE6DA} -6.06 & \cellcolor[HTML]{F6B394} -17.26 & \cellcolor[HTML]{E0ECF3} 6.24 & \cellcolor[HTML]{F9F0EB} -2.36 & \cellcolor[HTML]{F3F5F6} 1.05 & \cellcolor[HTML]{F0F4F6} 1.85 \\
American black bear & \cellcolor[HTML]{F7F5F4} -0.73 & \cellcolor[HTML]{F5F6F7} 0.62 & \cellcolor[HTML]{D4E6F1} 9.11 & \cellcolor[HTML]{EFF3F5} 2.16 & \cellcolor[HTML]{D1E5F0} 10.11 & \cellcolor[HTML]{FAE8DE} -5.47 & \cellcolor[HTML]{FCDECD} -8.63 & \cellcolor[HTML]{F5F6F7} 0.43 & \cellcolor[HTML]{F8F3F0} -1.46 & \cellcolor[HTML]{F9EEE7} -3.34 & \cellcolor[HTML]{A9D1E5} 16.15 \\
ashcan & \cellcolor[HTML]{F8F2EF} -1.57 & \cellcolor[HTML]{EAF1F5} 3.32 & \cellcolor[HTML]{EAF1F5} 3.30 & \cellcolor[HTML]{E3EDF3} 5.12 & \cellcolor[HTML]{EAF1F5} 3.39 & \cellcolor[HTML]{FDDCC9} -9.38 & \cellcolor[HTML]{F7B799} -16.62 & \cellcolor[HTML]{EAF1F5} 3.29 & \cellcolor[HTML]{F8F3F0} -1.52 & \cellcolor[HTML]{ECF2F5} 2.92 & \cellcolor[HTML]{EFF3F5} 1.99 \\
ballpoint & \cellcolor[HTML]{FBE3D4} -7.20 & \cellcolor[HTML]{E6EFF4} 4.64 & \cellcolor[HTML]{F2F5F6} 1.45 & \cellcolor[HTML]{CFE4EF} 10.28 & \cellcolor[HTML]{E3EDF3} 5.11 & \cellcolor[HTML]{F6B191} -17.88 & \cellcolor[HTML]{CF5246}\color[HTML]{F1F1F1}-31.99 & \cellcolor[HTML]{B6D7E8} 14.32 & \cellcolor[HTML]{FCDECD} -8.65 & \cellcolor[HTML]{CCE2EF} 10.67 & \cellcolor[HTML]{98C8E0} 18.84 \\
beach wagon & \cellcolor[HTML]{F9EFE9} -3.10 & \cellcolor[HTML]{F2F5F6} 1.26 & \cellcolor[HTML]{F0F4F6} 1.68 & \cellcolor[HTML]{E7F0F4} 3.99 & \cellcolor[HTML]{EDF2F5} 2.43 & \cellcolor[HTML]{FBE4D6} -6.88 & \cellcolor[HTML]{F9C6AC} -13.68 & \cellcolor[HTML]{EAF1F5} 3.40 & \cellcolor[HTML]{F9EDE5} -3.73 & \cellcolor[HTML]{F3F5F6} 0.81 & \cellcolor[HTML]{F2F5F6} 1.54 \\
boathouse & \cellcolor[HTML]{F9C2A7} -14.50 & \cellcolor[HTML]{B6D7E8} 14.26 & \cellcolor[HTML]{E1EDF3} 5.66 & \cellcolor[HTML]{96C7DF} 19.27 & \cellcolor[HTML]{D1E5F0} 9.97 & \cellcolor[HTML]{F6AF8E} -18.29 & \cellcolor[HTML]{E98B6E}\color[HTML]{F1F1F1}-23.63 & \cellcolor[HTML]{9DCBE1} 18.08 & \cellcolor[HTML]{FDDCC9} -9.59 & \cellcolor[HTML]{BBDAEA} 13.59 & \cellcolor[HTML]{CAE1EE} 11.24 \\
bullet train & \cellcolor[HTML]{E0ECF3} 6.15 & \cellcolor[HTML]{E9F0F4} 3.60 & \cellcolor[HTML]{E0ECF3} 6.14 & \cellcolor[HTML]{DDEBF2} 6.87 & \cellcolor[HTML]{CCE2EF} 10.62 & \cellcolor[HTML]{FACAB1} -13.04 & \cellcolor[HTML]{EF9979} -21.78 & \cellcolor[HTML]{F5F6F7} 0.78 & \cellcolor[HTML]{F5F6F7} 0.63 & \cellcolor[HTML]{FAE9DF} -4.96 & \cellcolor[HTML]{AED3E6} 15.59 \\
carbonara & \cellcolor[HTML]{F7F6F6} -0.03 & \cellcolor[HTML]{F7F5F4} -0.58 & \cellcolor[HTML]{FBE4D6} -6.90 & \cellcolor[HTML]{F8F3F0} -1.46 & \cellcolor[HTML]{FBE4D6} -6.66 & \cellcolor[HTML]{E1EDF3} 5.51 & \cellcolor[HTML]{F3F5F6} 1.02 & \cellcolor[HTML]{F8F3F0} -1.28 & \cellcolor[HTML]{EAF1F5} 3.21 & \cellcolor[HTML]{F2F5F6} 1.19 & \cellcolor[HTML]{E3EDF3} 5.32 \\
cellular telephone & \cellcolor[HTML]{F9EDE5} -3.52 & \cellcolor[HTML]{E1EDF3} 5.58 & \cellcolor[HTML]{EDF2F5} 2.47 & \cellcolor[HTML]{DDEBF2} 6.95 & \cellcolor[HTML]{E7F0F4} 4.26 & \cellcolor[HTML]{FBE3D4} -7.16 & \cellcolor[HTML]{F5AC8B} -18.59 & \cellcolor[HTML]{D7E8F1} 8.51 & \cellcolor[HTML]{FAE8DE} -5.45 & \cellcolor[HTML]{DDEBF2} 6.80 & \cellcolor[HTML]{E1EDF3} 5.71 \\
chest & \cellcolor[HTML]{F9EBE3} -4.23 & \cellcolor[HTML]{EFF3F5} 2.02 & \cellcolor[HTML]{EFF3F5} 2.30 & \cellcolor[HTML]{ECF2F5} 3.03 & \cellcolor[HTML]{E7F0F4} 4.00 & \cellcolor[HTML]{FBE5D8} -6.27 & \cellcolor[HTML]{FCDECD} -8.90 & \cellcolor[HTML]{EDF2F5} 2.37 & \cellcolor[HTML]{F8F1ED} -2.22 & \cellcolor[HTML]{ECF2F5} 2.81 & \cellcolor[HTML]{F0F4F6} 1.79 \\
clog & \cellcolor[HTML]{F7F5F4} -0.59 & \cellcolor[HTML]{E7F0F4} 4.30 & \cellcolor[HTML]{F5F6F7} 0.59 & \cellcolor[HTML]{F0F4F6} 1.67 & \cellcolor[HTML]{EDF2F5} 2.58 & \cellcolor[HTML]{FBE5D8} -6.54 & \cellcolor[HTML]{F8BFA4} -15.05 & \cellcolor[HTML]{EFF3F5} 2.34 & \cellcolor[HTML]{F7F5F4} -0.55 & \cellcolor[HTML]{EDF2F5} 2.48 & \cellcolor[HTML]{EFF3F5} 1.96 \\
container ship & \cellcolor[HTML]{FBE4D6} -6.68 & \cellcolor[HTML]{C5DFEC} 11.74 & \cellcolor[HTML]{F5F6F7} 0.43 & \cellcolor[HTML]{A2CDE3} 17.30 & \cellcolor[HTML]{D7E8F1} 8.45 & \cellcolor[HTML]{F9C2A7} -14.74 & \cellcolor[HTML]{E17860}\color[HTML]{F1F1F1}-26.55 & \cellcolor[HTML]{B6D7E8} 14.36 & \cellcolor[HTML]{F9EBE3} -4.27 & \cellcolor[HTML]{C2DDEC} 12.41 & \cellcolor[HTML]{93C6DE} 19.54 \\
digital watch & \cellcolor[HTML]{FAE7DC} -5.76 & \cellcolor[HTML]{E6EFF4} 4.31 & \cellcolor[HTML]{F6F7F7} 0.24 & \cellcolor[HTML]{E3EDF3} 5.45 & \cellcolor[HTML]{E7F0F4} 4.15 & \cellcolor[HTML]{FDDDCB} -9.10 & \cellcolor[HTML]{F3A481} -20.19 & \cellcolor[HTML]{E4EEF4} 4.76 & \cellcolor[HTML]{FAE7DC} -5.77 & \cellcolor[HTML]{E3EDF3} 5.35 & \cellcolor[HTML]{EAF1F5} 3.25 \\
dining table & \cellcolor[HTML]{FAEAE1} -4.56 & \cellcolor[HTML]{F7F5F4} -0.52 & \cellcolor[HTML]{E4EEF4} 4.79 & \cellcolor[HTML]{ECF2F5} 2.81 & \cellcolor[HTML]{BDDBEA} 13.17 & \cellcolor[HTML]{FDDCC9} -9.59 & \cellcolor[HTML]{FBE3D4} -7.35 & \cellcolor[HTML]{F8F2EF} -1.85 & \cellcolor[HTML]{FAE8DE} -5.17 & \cellcolor[HTML]{EFF3F5} 2.21 & \cellcolor[HTML]{F8F3F0} -1.49 \\
dog (kuvasz) & \cellcolor[HTML]{F3F5F6} 1.13 & \cellcolor[HTML]{F8F4F2} -0.92 & \cellcolor[HTML]{F7F6F6} -0.31 & \cellcolor[HTML]{EAF1F5} 3.50 & \cellcolor[HTML]{F0F4F6} 1.79 & \cellcolor[HTML]{F0F4F6} 1.78 & \cellcolor[HTML]{F9EDE5} -3.64 & \cellcolor[HTML]{F9EFE9} -2.97 & \cellcolor[HTML]{F0F4F6} 1.72 & \cellcolor[HTML]{FAE8DE} -5.20 & \cellcolor[HTML]{F7F6F6} -0.13 \\
giant panda & \cellcolor[HTML]{EFF3F5} 2.00 & \cellcolor[HTML]{F5F6F7} 0.41 & \cellcolor[HTML]{F5F6F7} 0.74 & \cellcolor[HTML]{F7F6F6} -0.20 & \cellcolor[HTML]{E4EEF4} 4.89 & \cellcolor[HTML]{F8F4F2} -1.13 & \cellcolor[HTML]{F9F0EB} -2.37 & \cellcolor[HTML]{F2F5F6} 1.47 & \cellcolor[HTML]{F7F6F6} -0.09 & \cellcolor[HTML]{F8F3F0} -1.51 & \cellcolor[HTML]{F5F6F7} 0.52 \\
gibbon & \cellcolor[HTML]{F7F6F6} -0.25 & \cellcolor[HTML]{F3F5F6} 1.10 & \cellcolor[HTML]{F5F6F7} 0.78 & \cellcolor[HTML]{EDF2F5} 2.45 & \cellcolor[HTML]{E4EEF4} 4.97 & \cellcolor[HTML]{FAE8DE} -5.17 & \cellcolor[HTML]{FCDECD} -8.93 & \cellcolor[HTML]{F5F6F7} 0.73 & \cellcolor[HTML]{F5F6F7} 0.41 & \cellcolor[HTML]{F2F5F6} 1.25 & \cellcolor[HTML]{E0ECF3} 5.88 \\
goldfinch & \cellcolor[HTML]{F6F7F7} 0.20 & \cellcolor[HTML]{E6EFF4} 4.57 & \cellcolor[HTML]{E6EFF4} 4.34 & \cellcolor[HTML]{E4EEF4} 4.78 & \cellcolor[HTML]{CAE1EE} 11.24 & \cellcolor[HTML]{FBCEB7} -12.28 & \cellcolor[HTML]{F7B99C} -16.04 & \cellcolor[HTML]{E1EDF3} 5.61 & \cellcolor[HTML]{F9EBE3} -4.00 & \cellcolor[HTML]{E9F0F4} 3.58 & \cellcolor[HTML]{DAE9F2} 7.58 \\
goldfish & \cellcolor[HTML]{F9EDE5} -3.66 & \cellcolor[HTML]{F2F5F6} 1.55 & \cellcolor[HTML]{F2F5F6} 1.55 & \cellcolor[HTML]{EAF1F5} 3.34 & \cellcolor[HTML]{DEEBF2} 6.47 & \cellcolor[HTML]{FBE5D8} -6.26 & \cellcolor[HTML]{F9C6AC} -14.02 & \cellcolor[HTML]{F9EDE5} -3.71 & \cellcolor[HTML]{F8F2EF} -1.65 & \cellcolor[HTML]{F8F4F2} -1.06 & \cellcolor[HTML]{F9EBE3} -4.26 \\
golf ball & \cellcolor[HTML]{F9EEE7} -3.51 & \cellcolor[HTML]{F3F5F6} 0.85 & \cellcolor[HTML]{F8F4F2} -0.80 & \cellcolor[HTML]{E1EDF3} 5.58 & \cellcolor[HTML]{DDEBF2} 6.92 & \cellcolor[HTML]{FCE2D2} -7.67 & \cellcolor[HTML]{F9C2A7} -14.84 & \cellcolor[HTML]{F0F4F6} 1.61 & \cellcolor[HTML]{FAE9DF} -4.71 & \cellcolor[HTML]{F9EFE9} -2.78 & \cellcolor[HTML]{CFE4EF} 10.25 \\
grand piano & \cellcolor[HTML]{FBE3D4} -7.42 & \cellcolor[HTML]{F0F4F6} 1.60 & \cellcolor[HTML]{DEEBF2} 6.28 & \cellcolor[HTML]{DEEBF2} 6.39 & \cellcolor[HTML]{E3EDF3} 5.11 & \cellcolor[HTML]{FAC8AF} -13.55 & \cellcolor[HTML]{E8896C}\color[HTML]{F1F1F1}-23.87 & \cellcolor[HTML]{E0ECF3} 6.14 & \cellcolor[HTML]{FBE6DA} -6.15 & \cellcolor[HTML]{ECF2F5} 3.00 & \cellcolor[HTML]{FBCEB7} -12.25 \\
hamster & \cellcolor[HTML]{F8F4F2} -0.96 & \cellcolor[HTML]{F8F4F2} -0.80 & \cellcolor[HTML]{F9EEE7} -3.20 & \cellcolor[HTML]{F7F5F4} -0.70 & \cellcolor[HTML]{F7F5F4} -0.45 & \cellcolor[HTML]{F0F4F6} 1.80 & \cellcolor[HTML]{F7F5F4} -0.51 & \cellcolor[HTML]{F6F7F7} 0.21 & \cellcolor[HTML]{F6F7F7} 0.14 & \cellcolor[HTML]{F3F5F6} 1.09 & \cellcolor[HTML]{F6F7F7} 0.16 \\
iron & \cellcolor[HTML]{FAEAE1} -4.30 & \cellcolor[HTML]{E9F0F4} 3.79 & \cellcolor[HTML]{EDF2F5} 2.41 & \cellcolor[HTML]{DDEBF2} 6.93 & \cellcolor[HTML]{E1EDF3} 5.85 & \cellcolor[HTML]{FDD9C4} -10.20 & \cellcolor[HTML]{F2A17F} -20.62 & \cellcolor[HTML]{DDEBF2} 6.72 & \cellcolor[HTML]{FAEAE1} -4.55 & \cellcolor[HTML]{E7F0F4} 4.09 & \cellcolor[HTML]{DEEBF2} 6.39 \\
lab coat & \cellcolor[HTML]{F7F6F6} -0.15 & \cellcolor[HTML]{EDF2F5} 2.62 & \cellcolor[HTML]{F7F5F4} -0.53 & \cellcolor[HTML]{E4EEF4} 4.86 & \cellcolor[HTML]{F9F0EB} -2.62 & \cellcolor[HTML]{F9F0EB} -2.36 & \cellcolor[HTML]{F7B99C} -16.23 & \cellcolor[HTML]{F0F4F6} 1.61 & \cellcolor[HTML]{F5F6F7} 0.56 & \cellcolor[HTML]{F9EEE7} -3.35 & \cellcolor[HTML]{F5F6F7} 0.68 \\
ladybug & \cellcolor[HTML]{F8F1ED} -2.20 & \cellcolor[HTML]{EAF1F5} 3.33 & \cellcolor[HTML]{EAF1F5} 3.37 & \cellcolor[HTML]{CAE1EE} 11.19 & \cellcolor[HTML]{9BC9E0} 18.42 & \cellcolor[HTML]{F5A886} -19.36 & \cellcolor[HTML]{D05548}\color[HTML]{F1F1F1}-31.38 & \cellcolor[HTML]{F7F6F6} -0.12 & \cellcolor[HTML]{FDDCC9} -9.54 & \cellcolor[HTML]{F3F5F6} 1.03 & \cellcolor[HTML]{CFE4EF} 10.22 \\
lemon & \cellcolor[HTML]{E9F0F4} 3.90 & \cellcolor[HTML]{F8F1ED} -2.07 & \cellcolor[HTML]{F7F6F6} -0.21 & \cellcolor[HTML]{F8F3F0} -1.55 & \cellcolor[HTML]{F5F6F7} 0.49 & \cellcolor[HTML]{F6F7F7} 0.32 & \cellcolor[HTML]{EDF2F5} 2.52 & \cellcolor[HTML]{F9F0EB} -2.65 & \cellcolor[HTML]{F7F5F4} -0.40 & \cellcolor[HTML]{F9EFE9} -3.04 & \cellcolor[HTML]{F7B799} -16.42 \\
mixing bowl & \cellcolor[HTML]{F6F7F7} 0.06 & \cellcolor[HTML]{F2F5F6} 1.24 & \cellcolor[HTML]{F7F5F4} -0.56 & \cellcolor[HTML]{F6F7F7} 0.39 & \cellcolor[HTML]{F8F4F2} -0.79 & \cellcolor[HTML]{F8F1ED} -2.12 & \cellcolor[HTML]{F9EFE9} -3.12 & \cellcolor[HTML]{F2F5F6} 1.23 & \cellcolor[HTML]{F6F7F7} 0.37 & \cellcolor[HTML]{F2F5F6} 1.20 & \cellcolor[HTML]{F8F4F2} -0.85 \\
motor scooter & \cellcolor[HTML]{F8F4F2} -1.06 & \cellcolor[HTML]{E6EFF4} 4.47 & \cellcolor[HTML]{EFF3F5} 2.29 & \cellcolor[HTML]{E3EDF3} 5.24 & \cellcolor[HTML]{E7F0F4} 4.21 & \cellcolor[HTML]{FBE4D6} -6.94 & \cellcolor[HTML]{E98B6E}\color[HTML]{F1F1F1}-23.61 & \cellcolor[HTML]{E3EDF3} 5.09 & \cellcolor[HTML]{F8F4F2} -1.12 & \cellcolor[HTML]{EFF3F5} 2.13 & \cellcolor[HTML]{FAEAE1} -4.60 \\
padlock & \cellcolor[HTML]{F9F0EB} -2.61 & \cellcolor[HTML]{E1EDF3} 5.85 & \cellcolor[HTML]{F2F5F6} 1.54 & \cellcolor[HTML]{D5E7F1} 8.78 & \cellcolor[HTML]{DEEBF2} 6.29 & \cellcolor[HTML]{FDDDCB} -9.29 & \cellcolor[HTML]{F7B799} -16.47 & \cellcolor[HTML]{E4EEF4} 5.01 & \cellcolor[HTML]{F8F4F2} -1.00 & \cellcolor[HTML]{E1EDF3} 5.53 & \cellcolor[HTML]{F9EDE5} -3.90 \\
park bench & \cellcolor[HTML]{FCD5BF} -11.21 & \cellcolor[HTML]{E0ECF3} 5.87 & \cellcolor[HTML]{E4EEF4} 5.03 & \cellcolor[HTML]{D4E6F1} 9.03 & \cellcolor[HTML]{DAE9F2} 7.67 & \cellcolor[HTML]{F5AC8B} -18.71 & \cellcolor[HTML]{EB9172}\color[HTML]{F1F1F1}-22.86 & \cellcolor[HTML]{D7E8F1} 8.39 & \cellcolor[HTML]{FBE3D4} -7.17 & \cellcolor[HTML]{E0ECF3} 5.89 & \cellcolor[HTML]{E7F0F4} 3.93 \\
purse & \cellcolor[HTML]{F8F3F0} -1.50 & \cellcolor[HTML]{EDF2F5} 2.51 & \cellcolor[HTML]{F9EFE9} -2.90 & \cellcolor[HTML]{EAF1F5} 3.47 & \cellcolor[HTML]{F5F6F7} 0.56 & \cellcolor[HTML]{F8F4F2} -0.88 & \cellcolor[HTML]{FBE3D4} -7.05 & \cellcolor[HTML]{ECF2F5} 3.07 & \cellcolor[HTML]{F2F5F6} 1.21 & \cellcolor[HTML]{E6EFF4} 4.31 & \cellcolor[HTML]{D4E6F1} 9.09 \\
red fox & \cellcolor[HTML]{F7F6F6} -0.27 & \cellcolor[HTML]{E9F0F4} 3.71 & \cellcolor[HTML]{F2F5F6} 1.49 & \cellcolor[HTML]{EAF1F5} 3.47 & \cellcolor[HTML]{E7F0F4} 4.12 & \cellcolor[HTML]{FDDBC7} -9.95 & \cellcolor[HTML]{F8BDA1} -15.43 & \cellcolor[HTML]{EDF2F5} 2.68 & \cellcolor[HTML]{F8F4F2} -0.91 & \cellcolor[HTML]{E1EDF3} 5.76 & \cellcolor[HTML]{D8E9F1} 8.08 \\
Siamese cat & \cellcolor[HTML]{F5F6F7} 0.66 & \cellcolor[HTML]{F8F4F2} -1.03 & \cellcolor[HTML]{F5F6F7} 0.51 & \cellcolor[HTML]{F7F6F6} -0.32 & \cellcolor[HTML]{F6F7F7} 0.30 & \cellcolor[HTML]{F0F4F6} 1.76 & \cellcolor[HTML]{F9EDE5} -3.78 & \cellcolor[HTML]{F9F0EB} -2.42 & \cellcolor[HTML]{F7F5F4} -0.56 & \cellcolor[HTML]{F9F0EB} -2.53 & \cellcolor[HTML]{F7F5F4} -0.55 \\
street sign & \cellcolor[HTML]{FCDFCF} -8.32 & \cellcolor[HTML]{D4E6F1} 9.05 & \cellcolor[HTML]{EDF2F5} 2.61 & \cellcolor[HTML]{D2E6F0} 9.51 & \cellcolor[HTML]{DEEBF2} 6.61 & \cellcolor[HTML]{FACAB1} -12.90 & \cellcolor[HTML]{F4A683} -19.70 & \cellcolor[HTML]{D7E8F1} 8.28 & \cellcolor[HTML]{FDDCC9} -9.70 & \cellcolor[HTML]{D7E8F1} 8.44 & \cellcolor[HTML]{FBE6DA} -6.09 \\
streetcar & \cellcolor[HTML]{F9EDE5} -3.89 & \cellcolor[HTML]{DAE9F2} 7.49 & \cellcolor[HTML]{E9F0F4} 3.87 & \cellcolor[HTML]{E0ECF3} 6.12 & \cellcolor[HTML]{E3EDF3} 5.13 & \cellcolor[HTML]{F7B799} -16.59 & \cellcolor[HTML]{E27B62}\color[HTML]{F1F1F1}-26.04 & \cellcolor[HTML]{BDDBEA} 13.12 & \cellcolor[HTML]{F9EBE3} -4.11 & \cellcolor[HTML]{DAE9F2} 7.75 & \cellcolor[HTML]{F9EDE5} -3.67 \\
sulphur butterfly & \cellcolor[HTML]{FAE9DF} -4.89 & \cellcolor[HTML]{F7F6F6} -0.04 & \cellcolor[HTML]{F5F6F7} 0.71 & \cellcolor[HTML]{E7F0F4} 4.13 & \cellcolor[HTML]{E1EDF3} 5.48 & \cellcolor[HTML]{FBE5D8} -6.45 & \cellcolor[HTML]{F7B99C} -16.25 & \cellcolor[HTML]{F3F5F6} 0.98 & \cellcolor[HTML]{FBE4D6} -6.84 & \cellcolor[HTML]{F9EEE7} -3.44 & \cellcolor[HTML]{DDEBF2} 6.92 \\
table lamp & \cellcolor[HTML]{F8F1ED} -2.19 & \cellcolor[HTML]{E3EDF3} 5.14 & \cellcolor[HTML]{F3F5F6} 1.13 & \cellcolor[HTML]{E0ECF3} 6.02 & \cellcolor[HTML]{E9F0F4} 3.75 & \cellcolor[HTML]{F9C4A9} -14.23 & \cellcolor[HTML]{F8BDA1} -15.44 & \cellcolor[HTML]{E1EDF3} 5.65 & \cellcolor[HTML]{F9EEE7} -3.13 & \cellcolor[HTML]{F0F4F6} 1.76 & \cellcolor[HTML]{FCE0D0} -7.82 \\
television & \cellcolor[HTML]{EFF3F5} 2.00 & \cellcolor[HTML]{EFF3F5} 2.06 & \cellcolor[HTML]{F2F5F6} 1.52 & \cellcolor[HTML]{E7F0F4} 4.03 & \cellcolor[HTML]{E3EDF3} 5.31 & \cellcolor[HTML]{F9EBE3} -4.08 & \cellcolor[HTML]{FDDBC7} -10.13 & \cellcolor[HTML]{EAF1F5} 3.19 & \cellcolor[HTML]{F8F2EF} -1.92 & \cellcolor[HTML]{E6EFF4} 4.36 & \cellcolor[HTML]{F9EDE5} -3.88 \\
tiger shark & \cellcolor[HTML]{F9F0EB} -2.66 & \cellcolor[HTML]{EAF1F5} 3.23 & \cellcolor[HTML]{EAF1F5} 3.22 & \cellcolor[HTML]{DDEBF2} 6.84 & \cellcolor[HTML]{E1EDF3} 5.66 & \cellcolor[HTML]{FBE6DA} -6.06 & \cellcolor[HTML]{F6B394} -17.23 & \cellcolor[HTML]{F0F4F6} 1.83 & \cellcolor[HTML]{F8F1ED} -2.22 & \cellcolor[HTML]{F7F5F4} -0.74 & \cellcolor[HTML]{9BC9E0} 18.55 \\
toilet seat & \cellcolor[HTML]{F3F5F6} 1.05 & \cellcolor[HTML]{ECF2F5} 3.04 & \cellcolor[HTML]{E6EFF4} 4.54 & \cellcolor[HTML]{E7F0F4} 4.08 & \cellcolor[HTML]{E0ECF3} 5.93 & \cellcolor[HTML]{FAE8DE} -5.25 & \cellcolor[HTML]{FCD5BF} -11.14 & \cellcolor[HTML]{E3EDF3} 5.34 & \cellcolor[HTML]{F8F4F2} -1.10 & \cellcolor[HTML]{F6F7F7} 0.12 & \cellcolor[HTML]{F9EFE9} -3.00 \\
tree frog & \cellcolor[HTML]{F7F6F6} -0.26 & \cellcolor[HTML]{E7F0F4} 4.15 & \cellcolor[HTML]{F9EBE3} -4.22 & \cellcolor[HTML]{E1EDF3} 5.48 & \cellcolor[HTML]{F2F5F6} 1.49 & \cellcolor[HTML]{FCE2D2} -7.43 & \cellcolor[HTML]{F9C2A7} -14.46 & \cellcolor[HTML]{EFF3F5} 2.20 & \cellcolor[HTML]{F8F4F2} -1.11 & \cellcolor[HTML]{E9F0F4} 3.81 & \cellcolor[HTML]{F0F4F6} 1.73 \\
umbrella & \cellcolor[HTML]{F9EDE5} -3.59 & \cellcolor[HTML]{EAF1F5} 3.41 & \cellcolor[HTML]{F8F4F2} -1.07 & \cellcolor[HTML]{E6EFF4} 4.67 & \cellcolor[HTML]{F5F6F7} 0.57 & \cellcolor[HTML]{F9EDE5} -3.64 & \cellcolor[HTML]{FBCCB4} -12.66 & \cellcolor[HTML]{EDF2F5} 2.44 & \cellcolor[HTML]{F9EFE9} -3.07 & \cellcolor[HTML]{F7F5F4} -0.49 & \cellcolor[HTML]{F9C6AC} -13.84 \\
vase & \cellcolor[HTML]{F9EEE7} -3.38 & \cellcolor[HTML]{DEEBF2} 6.34 & \cellcolor[HTML]{F9EEE7} -3.50 & \cellcolor[HTML]{E1EDF3} 5.52 & \cellcolor[HTML]{F5F6F7} 0.52 & \cellcolor[HTML]{FDD9C4} -10.40 & \cellcolor[HTML]{EC9374}\color[HTML]{F1F1F1}-22.52 & \cellcolor[HTML]{E9F0F4} 3.65 & \cellcolor[HTML]{F2F5F6} 1.34 & \cellcolor[HTML]{ECF2F5} 2.98 & \cellcolor[HTML]{FDDCC9} -9.59 \\
water bottle & \cellcolor[HTML]{F8F4F2} -1.05 & \cellcolor[HTML]{F5F6F7} 0.52 & \cellcolor[HTML]{F9EEE7} -3.20 & \cellcolor[HTML]{EFF3F5} 2.33 & \cellcolor[HTML]{F8F3F0} -1.32 & \cellcolor[HTML]{FCE2D2} -7.70 & \cellcolor[HTML]{F3A481} -20.06 & \cellcolor[HTML]{E1EDF3} 5.53 & \cellcolor[HTML]{F7F5F4} -0.40 & \cellcolor[HTML]{EAF1F5} 3.15 & \cellcolor[HTML]{FACAB1} -13.07 \\
water tower & \cellcolor[HTML]{FCDFCF} -8.22 & \cellcolor[HTML]{DBEAF2} 7.35 & \cellcolor[HTML]{ECF2F5} 3.10 & \cellcolor[HTML]{D4E6F1} 9.19 & \cellcolor[HTML]{EFF3F5} 2.17 & \cellcolor[HTML]{F4A683} -19.73 & \cellcolor[HTML]{E17860}\color[HTML]{F1F1F1}-26.49 & \cellcolor[HTML]{CCE2EF} 10.61 & \cellcolor[HTML]{F9EBE3} -3.92 & \cellcolor[HTML]{D8E9F1} 8.08 & \cellcolor[HTML]{CFE4EF} 10.31 \\
wild boar & \cellcolor[HTML]{F7F6F6} -0.13 & \cellcolor[HTML]{F0F4F6} 1.68 & \cellcolor[HTML]{F0F4F6} 1.86 & \cellcolor[HTML]{E0ECF3} 6.12 & \cellcolor[HTML]{E3EDF3} 5.13 & \cellcolor[HTML]{FBE4D6} -6.90 & \cellcolor[HTML]{FBCEB7} -12.30 & \cellcolor[HTML]{ECF2F5} 2.77 & \cellcolor[HTML]{F5F6F7} 0.74 & \cellcolor[HTML]{F2F5F6} 1.43 & \cellcolor[HTML]{CCE2EF} 10.91 \\
wood rabbit & \cellcolor[HTML]{F9EEE7} -3.21 & \cellcolor[HTML]{EDF2F5} 2.40 & \cellcolor[HTML]{F2F5F6} 1.20 & \cellcolor[HTML]{E0ECF3} 6.23 & \cellcolor[HTML]{DEEBF2} 6.37 & \cellcolor[HTML]{FBE3D4} -7.17 & \cellcolor[HTML]{FAC8AF} -13.62 & \cellcolor[HTML]{F7F6F6} -0.08 & \cellcolor[HTML]{F9F0EB} -2.72 & \cellcolor[HTML]{F0F4F6} 1.90 & \cellcolor[HTML]{D2E6F0} 9.41 \\
yawl & \cellcolor[HTML]{FDDDCB} -9.05 & \cellcolor[HTML]{DAE9F2} 7.79 & \cellcolor[HTML]{ECF2F5} 3.12 & \cellcolor[HTML]{D4E6F1} 9.13 & \cellcolor[HTML]{DEEBF2} 6.41 & \cellcolor[HTML]{F2A17F} -20.47 & \cellcolor[HTML]{DF765E}\color[HTML]{F1F1F1}-26.81 & \cellcolor[HTML]{BBDAEA} 13.41 & \cellcolor[HTML]{FAE7DC} -5.82 & \cellcolor[HTML]{DAE9F2} 7.56 & \cellcolor[HTML]{F8F4F2} -0.79 \\
\bottomrule
\end{tabular}
}
\end{table}

\begin{table}[tbp]
\scriptsize
\sisetup{table-format=+2.1}

\caption{\textbf{Background bias (ImageNetS50 concepts, GloCE):} The values show in \% how much the average test IoU of vanilla CEs increases (marked \emph{red}) respectively decreases (marked \emph{blue}) on specific background categories compared to performance on arbitrary backgrounds. Test samples are created via simple background pasting (4 random variants per test sample). CEs are globalized LoCEs for the shown ImageNetS50 concepts and late model layers, IoUs averaged over 7 models.
}\label{tab:iou-per-bg-cat-imagenet-gloce}
\renewcommand\arraystretch{1.2}
\newcommand{\tabh}[1]{\multicolumn{1}{c}{\parbox[t]{3.2em}{\centering\sffamily #1}}}

\scriptsize
\resizebox{.98\linewidth}{!}{%
  \color{black}%
\begin{tabular}{@{}>{\sffamily}lSSSSSSSSSSS@{}}
VOC Concept & \tabh{arch.} & \tabh{at water} & \tabh{botanical} & \tabh{field} & \tabh{forest} & \tabh{indoors} & \tabh{mach.} & \tabh{open l.} & \tabh{road} & \tabh{snow} & \tabh{vanilla} \\\midrule[\heavyrulewidth]
African elephant & \cellcolor[HTML]{FBE3D4} -7.24 & \cellcolor[HTML]{F2F5F6} 1.37 & \cellcolor[HTML]{F3F5F6} 0.89 & \cellcolor[HTML]{E3EDF3} 5.36 & \cellcolor[HTML]{ECF2F5} 2.90 & \cellcolor[HTML]{FCD3BC} -11.71 & \cellcolor[HTML]{F6B394} -17.22 & \cellcolor[HTML]{EDF2F5} 2.36 & \cellcolor[HTML]{F9EFE9} -2.81 & \cellcolor[HTML]{F8F3F0} -1.38 & \cellcolor[HTML]{B6D7E8} 14.26 \\
agaric & \cellcolor[HTML]{F8F1ED} -2.28 & \cellcolor[HTML]{F2F5F6} 1.26 & \cellcolor[HTML]{F5F6F7} 0.62 & \cellcolor[HTML]{F3F5F6} 0.78 & \cellcolor[HTML]{E6EFF4} 4.59 & \cellcolor[HTML]{FCD7C2} -10.91 & \cellcolor[HTML]{FDDDCB} -9.36 & \cellcolor[HTML]{EFF3F5} 1.99 & \cellcolor[HTML]{F9EEE7} -3.29 & \cellcolor[HTML]{F7F5F4} -0.47 & \cellcolor[HTML]{E0ECF3} 5.88 \\
airliner & \cellcolor[HTML]{F8F2EF} -1.84 & \cellcolor[HTML]{E1EDF3} 5.51 & \cellcolor[HTML]{F2F5F6} 1.19 & \cellcolor[HTML]{E0ECF3} 6.00 & \cellcolor[HTML]{E7F0F4} 4.22 & \cellcolor[HTML]{FDDCC9} -9.47 & \cellcolor[HTML]{F19E7D} -20.92 & \cellcolor[HTML]{DBEAF2} 7.13 & \cellcolor[HTML]{F7F5F4} -0.65 & \cellcolor[HTML]{E6EFF4} 4.64 & \cellcolor[HTML]{B3D6E8} 14.79 \\
American black bear & \cellcolor[HTML]{F8F4F2} -0.84 & \cellcolor[HTML]{F3F5F6} 1.12 & \cellcolor[HTML]{DDEBF2} 7.02 & \cellcolor[HTML]{F0F4F6} 1.91 & \cellcolor[HTML]{D5E7F1} 8.82 & \cellcolor[HTML]{FAEAE1} -4.50 & \cellcolor[HTML]{FAE7DC} -5.61 & \cellcolor[HTML]{F0F4F6} 1.82 & \cellcolor[HTML]{F8F2EF} -1.95 & \cellcolor[HTML]{F8F3F0} -1.34 & \cellcolor[HTML]{8AC0DB} 20.81 \\
ashcan & \cellcolor[HTML]{F9EDE5} -3.58 & \cellcolor[HTML]{E4EEF4} 4.79 & \cellcolor[HTML]{EFF3F5} 2.02 & \cellcolor[HTML]{DDEBF2} 6.82 & \cellcolor[HTML]{E6EFF4} 4.66 & \cellcolor[HTML]{FAC8AF} -13.45 & \cellcolor[HTML]{F09C7B} -21.24 & \cellcolor[HTML]{E1EDF3} 5.61 & \cellcolor[HTML]{F9EFE9} -2.89 & \cellcolor[HTML]{E3EDF3} 5.44 & \cellcolor[HTML]{F9EFE9} -3.12 \\
ballpoint & \cellcolor[HTML]{FAE9DF} -4.99 & \cellcolor[HTML]{E0ECF3} 6.21 & \cellcolor[HTML]{F3F5F6} 1.17 & \cellcolor[HTML]{DBEAF2} 7.34 & \cellcolor[HTML]{E1EDF3} 5.78 & \cellcolor[HTML]{F7B596} -17.18 & \cellcolor[HTML]{DA6853}\color[HTML]{F1F1F1}-28.86 & \cellcolor[HTML]{B6D7E8} 14.23 & \cellcolor[HTML]{FDDCC9} -9.76 & \cellcolor[HTML]{D8E9F1} 7.81 & \cellcolor[HTML]{C5DFEC} 11.99 \\
beach wagon & \cellcolor[HTML]{F9EFE9} -3.07 & \cellcolor[HTML]{F0F4F6} 1.65 & \cellcolor[HTML]{F2F5F6} 1.37 & \cellcolor[HTML]{E9F0F4} 3.76 & \cellcolor[HTML]{EFF3F5} 2.15 & \cellcolor[HTML]{FAE9DF} -4.80 & \cellcolor[HTML]{FBCCB4} -12.74 & \cellcolor[HTML]{EAF1F5} 3.19 & \cellcolor[HTML]{F9EFE9} -2.87 & \cellcolor[HTML]{F2F5F6} 1.23 & \cellcolor[HTML]{EFF3F5} 2.26 \\
boathouse & \cellcolor[HTML]{EE9677} -22.17 & \cellcolor[HTML]{90C4DD} 20.00 & \cellcolor[HTML]{CAE1EE} 11.11 & \cellcolor[HTML]{6EAED2}\color[HTML]{F1F1F1}24.24 & \cellcolor[HTML]{9BC9E0} 18.61 & \cellcolor[HTML]{BF3338}\color[HTML]{F1F1F1}-35.95 & \cellcolor[HTML]{B61F2E}\color[HTML]{F1F1F1}-38.94 & \cellcolor[HTML]{78B4D5} 23.14 & \cellcolor[HTML]{FACAB1} -13.19 & \cellcolor[HTML]{B1D5E7} 14.99 & \cellcolor[HTML]{8DC2DC} 20.42 \\
bullet train & \cellcolor[HTML]{EFF3F5} 2.19 & \cellcolor[HTML]{F9F0EB} -2.64 & \cellcolor[HTML]{F8F4F2} -0.81 & \cellcolor[HTML]{F6F7F7} 0.01 & \cellcolor[HTML]{E3EDF3} 5.33 & \cellcolor[HTML]{FCDFCF} -8.41 & \cellcolor[HTML]{FACAB1} -12.90 & \cellcolor[HTML]{F8F2EF} -1.77 & \cellcolor[HTML]{EFF3F5} 2.03 & \cellcolor[HTML]{FBE5D8} -6.39 & \cellcolor[HTML]{87BEDA} 21.35 \\
carbonara & \cellcolor[HTML]{F8F3F0} -1.20 & \cellcolor[HTML]{F5F6F7} 0.74 & \cellcolor[HTML]{FAE7DC} -5.74 & \cellcolor[HTML]{F2F5F6} 1.46 & \cellcolor[HTML]{FBE6DA} -5.98 & \cellcolor[HTML]{EFF3F5} 2.09 & \cellcolor[HTML]{F9EEE7} -3.34 & \cellcolor[HTML]{F6F7F7} 0.32 & \cellcolor[HTML]{EFF3F5} 2.05 & \cellcolor[HTML]{F2F5F6} 1.55 & \cellcolor[HTML]{E9F0F4} 3.59 \\
cellular telephone & \cellcolor[HTML]{F9EDE5} -3.56 & \cellcolor[HTML]{D7E8F1} 8.38 & \cellcolor[HTML]{EDF2F5} 2.55 & \cellcolor[HTML]{D2E6F0} 9.57 & \cellcolor[HTML]{D8E9F1} 7.82 & \cellcolor[HTML]{FDD9C4} -10.31 & \cellcolor[HTML]{F6AF8E} -18.08 & \cellcolor[HTML]{C5DFEC} 12.06 & \cellcolor[HTML]{F9EEE7} -3.40 & \cellcolor[HTML]{D5E7F1} 8.96 & \cellcolor[HTML]{B8D8E9} 13.79 \\
chest & \cellcolor[HTML]{F9EFE9} -3.07 & \cellcolor[HTML]{F0F4F6} 1.82 & \cellcolor[HTML]{F0F4F6} 1.79 & \cellcolor[HTML]{EFF3F5} 2.17 & \cellcolor[HTML]{EDF2F5} 2.71 & \cellcolor[HTML]{FAE8DE} -5.14 & \cellcolor[HTML]{FBE4D6} -7.00 & \cellcolor[HTML]{EFF3F5} 1.97 & \cellcolor[HTML]{F8F3F0} -1.42 & \cellcolor[HTML]{ECF2F5} 2.91 & \cellcolor[HTML]{ECF2F5} 3.10 \\
clog & \cellcolor[HTML]{F7F5F4} -0.49 & \cellcolor[HTML]{E4EEF4} 5.03 & \cellcolor[HTML]{EDF2F5} 2.57 & \cellcolor[HTML]{EDF2F5} 2.65 & \cellcolor[HTML]{E3EDF3} 5.12 & \cellcolor[HTML]{FBE3D4} -7.18 & \cellcolor[HTML]{FCD5BF} -11.19 & \cellcolor[HTML]{EAF1F5} 3.16 & \cellcolor[HTML]{F9EFE9} -3.00 & \cellcolor[HTML]{F0F4F6} 1.82 & \cellcolor[HTML]{CFE4EF} 10.20 \\
container ship & \cellcolor[HTML]{FCDFCF} -8.47 & \cellcolor[HTML]{AED3E6} 15.59 & \cellcolor[HTML]{F2F5F6} 1.39 & \cellcolor[HTML]{B1D5E7} 14.94 & \cellcolor[HTML]{D4E6F1} 9.11 & \cellcolor[HTML]{F9C2A7} -14.82 & \cellcolor[HTML]{E58368}\color[HTML]{F1F1F1}-24.93 & \cellcolor[HTML]{BBDAEA} 13.42 & \cellcolor[HTML]{F8F3F0} -1.27 & \cellcolor[HTML]{B8D8E9} 13.89 & \cellcolor[HTML]{1A5899}\color[HTML]{F1F1F1}42.57 \\
digital watch & \cellcolor[HTML]{FBE3D4} -7.08 & \cellcolor[HTML]{E9F0F4} 3.66 & \cellcolor[HTML]{F5F6F7} 0.61 & \cellcolor[HTML]{E1EDF3} 5.59 & \cellcolor[HTML]{E6EFF4} 4.42 & \cellcolor[HTML]{FCDFCF} -8.35 & \cellcolor[HTML]{F5AC8B} -18.43 & \cellcolor[HTML]{E6EFF4} 4.62 & \cellcolor[HTML]{FAE9DF} -4.88 & \cellcolor[HTML]{DEEBF2} 6.27 & \cellcolor[HTML]{DEEBF2} 6.64 \\
dining table & \cellcolor[HTML]{FDDCC9} -9.56 & \cellcolor[HTML]{F7F6F6} -0.03 & \cellcolor[HTML]{D8E9F1} 8.05 & \cellcolor[HTML]{E4EEF4} 4.85 & \cellcolor[HTML]{8DC2DC} 20.64 & \cellcolor[HTML]{FDD9C4} -10.50 & \cellcolor[HTML]{FAEAE1} -4.53 & \cellcolor[HTML]{E9F0F4} 3.65 & \cellcolor[HTML]{FAEAE1} -4.68 & \cellcolor[HTML]{E3EDF3} 5.13 & \cellcolor[HTML]{246AAE}\color[HTML]{F1F1F1}39.02 \\
dog (kuvasz) & \cellcolor[HTML]{F6F7F7} 0.08 & \cellcolor[HTML]{F6F7F7} 0.10 & \cellcolor[HTML]{F3F5F6} 1.11 & \cellcolor[HTML]{E7F0F4} 4.01 & \cellcolor[HTML]{EDF2F5} 2.51 & \cellcolor[HTML]{FAEAE1} -4.42 & \cellcolor[HTML]{FAE8DE} -5.20 & \cellcolor[HTML]{F9F0EB} -2.73 & \cellcolor[HTML]{F5F6F7} 0.50 & \cellcolor[HTML]{FAE8DE} -5.12 & \cellcolor[HTML]{EFF3F5} 2.06 \\
giant panda & \cellcolor[HTML]{F0F4F6} 1.75 & \cellcolor[HTML]{F6F7F7} 0.12 & \cellcolor[HTML]{EDF2F5} 2.52 & \cellcolor[HTML]{F6F7F7} 0.32 & \cellcolor[HTML]{E4EEF4} 5.01 & \cellcolor[HTML]{F8F3F0} -1.40 & \cellcolor[HTML]{F9EDE5} -3.90 & \cellcolor[HTML]{F2F5F6} 1.37 & \cellcolor[HTML]{F8F3F0} -1.27 & \cellcolor[HTML]{F9F0EB} -2.46 & \cellcolor[HTML]{E9F0F4} 3.80 \\
gibbon & \cellcolor[HTML]{F8F3F0} -1.56 & \cellcolor[HTML]{F6F7F7} 0.27 & \cellcolor[HTML]{F8F4F2} -0.95 & \cellcolor[HTML]{F3F5F6} 1.12 & \cellcolor[HTML]{EAF1F5} 3.20 & \cellcolor[HTML]{F9EDE5} -3.86 & \cellcolor[HTML]{FBE3D4} -7.15 & \cellcolor[HTML]{F7F5F4} -0.59 & \cellcolor[HTML]{F8F4F2} -1.06 & \cellcolor[HTML]{ECF2F5} 2.86 & \cellcolor[HTML]{DBEAF2} 7.26 \\
goldfinch & \cellcolor[HTML]{EDF2F5} 2.47 & \cellcolor[HTML]{E9F0F4} 3.58 & \cellcolor[HTML]{F0F4F6} 1.76 & \cellcolor[HTML]{E7F0F4} 4.03 & \cellcolor[HTML]{DBEAF2} 7.34 & \cellcolor[HTML]{FDDDCB} -9.12 & \cellcolor[HTML]{FAE7DC} -5.78 & \cellcolor[HTML]{EDF2F5} 2.67 & \cellcolor[HTML]{F9F0EB} -2.66 & \cellcolor[HTML]{EDF2F5} 2.69 & \cellcolor[HTML]{C2DDEC} 12.22 \\
goldfish & \cellcolor[HTML]{F9F0EB} -2.45 & \cellcolor[HTML]{E7F0F4} 4.12 & \cellcolor[HTML]{F5F6F7} 0.64 & \cellcolor[HTML]{E6EFF4} 4.40 & \cellcolor[HTML]{E1EDF3} 5.75 & \cellcolor[HTML]{F8BB9E} -15.78 & \cellcolor[HTML]{FBCCB4} -12.69 & \cellcolor[HTML]{F5F6F7} 0.76 & \cellcolor[HTML]{F9EDE5} -3.84 & \cellcolor[HTML]{F3F5F6} 0.84 & \cellcolor[HTML]{F9EDE5} -3.71 \\
golf ball & \cellcolor[HTML]{FAEAE1} -4.31 & \cellcolor[HTML]{F0F4F6} 1.92 & \cellcolor[HTML]{F8F3F0} -1.56 & \cellcolor[HTML]{E3EDF3} 5.12 & \cellcolor[HTML]{EAF1F5} 3.41 & \cellcolor[HTML]{FCD3BC} -11.61 & \cellcolor[HTML]{FCD7C2} -10.58 & \cellcolor[HTML]{E9F0F4} 3.86 & \cellcolor[HTML]{FBE6DA} -5.87 & \cellcolor[HTML]{F8F1ED} -2.14 & \cellcolor[HTML]{BBDAEA} 13.29 \\
grand piano & \cellcolor[HTML]{FCDFCF} -8.48 & \cellcolor[HTML]{F0F4F6} 1.62 & \cellcolor[HTML]{EAF1F5} 3.42 & \cellcolor[HTML]{E7F0F4} 4.10 & \cellcolor[HTML]{E9F0F4} 3.89 & \cellcolor[HTML]{FCE2D2} -7.43 & \cellcolor[HTML]{F7B799} -16.73 & \cellcolor[HTML]{E1EDF3} 5.70 & \cellcolor[HTML]{FAEAE1} -4.49 & \cellcolor[HTML]{E6EFF4} 4.52 & \cellcolor[HTML]{FCD5BF} -11.17 \\
hamster & \cellcolor[HTML]{F8F3F0} -1.35 & \cellcolor[HTML]{F8F3F0} -1.49 & \cellcolor[HTML]{F9F0EB} -2.52 & \cellcolor[HTML]{F8F3F0} -1.46 & \cellcolor[HTML]{F8F2EF} -1.69 & \cellcolor[HTML]{F0F4F6} 1.66 & \cellcolor[HTML]{F0F4F6} 1.80 & \cellcolor[HTML]{F8F1ED} -2.21 & \cellcolor[HTML]{F6F7F7} 0.02 & \cellcolor[HTML]{F7F6F6} -0.26 & \cellcolor[HTML]{B8D8E9} 13.88 \\
iron & \cellcolor[HTML]{FAE8DE} -5.37 & \cellcolor[HTML]{DEEBF2} 6.58 & \cellcolor[HTML]{EDF2F5} 2.70 & \cellcolor[HTML]{D8E9F1} 8.06 & \cellcolor[HTML]{E0ECF3} 5.93 & \cellcolor[HTML]{FAC8AF} -13.45 & \cellcolor[HTML]{F7B99C} -16.15 & \cellcolor[HTML]{D2E6F0} 9.56 & \cellcolor[HTML]{F9EFE9} -2.75 & \cellcolor[HTML]{DEEBF2} 6.52 & \cellcolor[HTML]{F7F5F4} -0.57 \\
lab coat & \cellcolor[HTML]{F7F5F4} -0.67 & \cellcolor[HTML]{F5F6F7} 0.45 & \cellcolor[HTML]{F7F6F6} -0.01 & \cellcolor[HTML]{F0F4F6} 1.66 & \cellcolor[HTML]{F8F4F2} -0.92 & \cellcolor[HTML]{F8F2EF} -1.68 & \cellcolor[HTML]{FCDECD} -8.97 & \cellcolor[HTML]{F7F6F6} -0.02 & \cellcolor[HTML]{F7F6F6} -0.31 & \cellcolor[HTML]{FAE8DE} -5.09 & \cellcolor[HTML]{F6F7F7} 0.21 \\
ladybug & \cellcolor[HTML]{F9F0EB} -2.66 & \cellcolor[HTML]{E0ECF3} 6.25 & \cellcolor[HTML]{F9F0EB} -2.39 & \cellcolor[HTML]{D7E8F1} 8.27 & \cellcolor[HTML]{C2DDEC} 12.38 & \cellcolor[HTML]{EE9677} -22.10 & \cellcolor[HTML]{F3A481} -20.22 & \cellcolor[HTML]{D2E6F0} 9.64 & \cellcolor[HTML]{FCE2D2} -7.76 & \cellcolor[HTML]{D8E9F1} 7.93 & \cellcolor[HTML]{AED3E6} 15.50 \\
lemon & \cellcolor[HTML]{F3F5F6} 0.79 & \cellcolor[HTML]{F6F7F7} 0.24 & \cellcolor[HTML]{F8F4F2} -0.89 & \cellcolor[HTML]{F7F6F6} -0.08 & \cellcolor[HTML]{F2F5F6} 1.21 & \cellcolor[HTML]{F8F1ED} -2.25 & \cellcolor[HTML]{F5F6F7} 0.59 & \cellcolor[HTML]{F6F7F7} 0.36 & \cellcolor[HTML]{F8F1ED} -2.27 & \cellcolor[HTML]{F3F5F6} 1.13 & \cellcolor[HTML]{F5A886} -19.31 \\
mixing bowl & \cellcolor[HTML]{F5F6F7} 0.52 & \cellcolor[HTML]{F6F7F7} 0.35 & \cellcolor[HTML]{F8F4F2} -0.85 & \cellcolor[HTML]{F6F7F7} 0.05 & \cellcolor[HTML]{F8F4F2} -0.81 & \cellcolor[HTML]{F8F1ED} -2.21 & \cellcolor[HTML]{F8F2EF} -1.86 & \cellcolor[HTML]{F7F6F6} -0.01 & \cellcolor[HTML]{F6F7F7} 0.28 & \cellcolor[HTML]{F3F5F6} 1.04 & \cellcolor[HTML]{F3F5F6} 1.15 \\
motor scooter & \cellcolor[HTML]{F7F6F6} -0.28 & \cellcolor[HTML]{EAF1F5} 3.24 & \cellcolor[HTML]{F3F5F6} 1.04 & \cellcolor[HTML]{E6EFF4} 4.35 & \cellcolor[HTML]{E9F0F4} 3.58 & \cellcolor[HTML]{FAE9DF} -4.81 & \cellcolor[HTML]{F19E7D} -20.80 & \cellcolor[HTML]{E7F0F4} 4.00 & \cellcolor[HTML]{F7F5F4} -0.48 & \cellcolor[HTML]{F6F7F7} 0.22 & \cellcolor[HTML]{F9EEE7} -3.48 \\
padlock & \cellcolor[HTML]{FCDFCF} -8.23 & \cellcolor[HTML]{E4EEF4} 4.70 & \cellcolor[HTML]{EAF1F5} 3.32 & \cellcolor[HTML]{E0ECF3} 6.19 & \cellcolor[HTML]{B8D8E9} 13.74 & \cellcolor[HTML]{F6AF8E} -18.22 & \cellcolor[HTML]{FCDFCF} -8.38 & \cellcolor[HTML]{D5E7F1} 8.89 & \cellcolor[HTML]{F8F1ED} -2.29 & \cellcolor[HTML]{DAE9F2} 7.59 & \cellcolor[HTML]{FAEAE1} -4.61 \\
park bench & \cellcolor[HTML]{FBCEB7} -12.42 & \cellcolor[HTML]{D5E7F1} 8.60 & \cellcolor[HTML]{D5E7F1} 8.85 & \cellcolor[HTML]{C0DCEB} 12.61 & \cellcolor[HTML]{C5DFEC} 11.88 & \cellcolor[HTML]{EF9979} -21.73 & \cellcolor[HTML]{E37E64}\color[HTML]{F1F1F1}-25.62 & \cellcolor[HTML]{D2E6F0} 9.55 & \cellcolor[HTML]{FBE3D4} -7.14 & \cellcolor[HTML]{E0ECF3} 6.03 & \cellcolor[HTML]{ACD2E5} 15.86 \\
purse & \cellcolor[HTML]{F8F1ED} -2.01 & \cellcolor[HTML]{ECF2F5} 3.12 & \cellcolor[HTML]{F9EFE9} -2.92 & \cellcolor[HTML]{E9F0F4} 3.85 & \cellcolor[HTML]{F8F4F2} -0.83 & \cellcolor[HTML]{F8F2EF} -1.87 & \cellcolor[HTML]{FDDCC9} -9.60 & \cellcolor[HTML]{ECF2F5} 3.02 & \cellcolor[HTML]{F5F6F7} 0.57 & \cellcolor[HTML]{E7F0F4} 4.06 & \cellcolor[HTML]{DDEBF2} 6.76 \\
red fox & \cellcolor[HTML]{F9EFE9} -3.10 & \cellcolor[HTML]{E0ECF3} 6.05 & \cellcolor[HTML]{E9F0F4} 3.89 & \cellcolor[HTML]{E4EEF4} 4.79 & \cellcolor[HTML]{DBEAF2} 7.14 & \cellcolor[HTML]{FBCCB4} -12.69 & \cellcolor[HTML]{F8BB9E} -15.84 & \cellcolor[HTML]{E7F0F4} 4.07 & \cellcolor[HTML]{F9F0EB} -2.42 & \cellcolor[HTML]{D5E7F1} 8.90 & \cellcolor[HTML]{8AC0DB} 21.08 \\
Siamese cat & \cellcolor[HTML]{F3F5F6} 0.94 & \cellcolor[HTML]{F7F5F4} -0.55 & \cellcolor[HTML]{F7F5F4} -0.63 & \cellcolor[HTML]{F6F7F7} 0.32 & \cellcolor[HTML]{F6F7F7} 0.03 & \cellcolor[HTML]{EDF2F5} 2.53 & \cellcolor[HTML]{F8F1ED} -2.23 & \cellcolor[HTML]{F9F0EB} -2.45 & \cellcolor[HTML]{F7F5F4} -0.71 & \cellcolor[HTML]{F8F1ED} -2.26 & \cellcolor[HTML]{DDEBF2} 6.85 \\
street sign & \cellcolor[HTML]{F9F0EB} -2.41 & \cellcolor[HTML]{EFF3F5} 2.34 & \cellcolor[HTML]{E7F0F4} 4.19 & \cellcolor[HTML]{E9F0F4} 3.65 & \cellcolor[HTML]{DDEBF2} 6.67 & \cellcolor[HTML]{FCDECD} -8.84 & \cellcolor[HTML]{FCD5BF} -11.30 & \cellcolor[HTML]{E9F0F4} 3.67 & \cellcolor[HTML]{FAE8DE} -5.28 & \cellcolor[HTML]{F8F4F2} -0.95 & \cellcolor[HTML]{F9EBE3} -4.13 \\
streetcar & \cellcolor[HTML]{F8F1ED} -2.30 & \cellcolor[HTML]{E6EFF4} 4.50 & \cellcolor[HTML]{E3EDF3} 5.11 & \cellcolor[HTML]{E1EDF3} 5.50 & \cellcolor[HTML]{E3EDF3} 5.25 & \cellcolor[HTML]{FCD7C2} -10.65 & \cellcolor[HTML]{F6AF8E} -18.14 & \cellcolor[HTML]{DAE9F2} 7.57 & \cellcolor[HTML]{F7F6F6} -0.20 & \cellcolor[HTML]{EDF2F5} 2.59 & \cellcolor[HTML]{D7E8F1} 8.26 \\
sulphur butterfly & \cellcolor[HTML]{F9EFE9} -3.04 & \cellcolor[HTML]{F6F7F7} 0.37 & \cellcolor[HTML]{F8F4F2} -1.10 & \cellcolor[HTML]{EDF2F5} 2.37 & \cellcolor[HTML]{ECF2F5} 3.11 & \cellcolor[HTML]{FBD0B9} -12.05 & \cellcolor[HTML]{FCDECD} -8.94 & \cellcolor[HTML]{EFF3F5} 2.13 & \cellcolor[HTML]{F9EBE3} -3.94 & \cellcolor[HTML]{F3F5F6} 1.05 & \cellcolor[HTML]{F2F5F6} 1.56 \\
table lamp & \cellcolor[HTML]{F8F2EF} -1.87 & \cellcolor[HTML]{E7F0F4} 4.13 & \cellcolor[HTML]{F8F2EF} -1.64 & \cellcolor[HTML]{F0F4F6} 1.71 & \cellcolor[HTML]{EDF2F5} 2.60 & \cellcolor[HTML]{FCE0D0} -7.92 & \cellcolor[HTML]{FCD5BF} -11.15 & \cellcolor[HTML]{E4EEF4} 4.74 & \cellcolor[HTML]{F9EEE7} -3.37 & \cellcolor[HTML]{EAF1F5} 3.28 & \cellcolor[HTML]{FAE7DC} -5.66 \\
television & \cellcolor[HTML]{F2F5F6} 1.19 & \cellcolor[HTML]{F3F5F6} 0.94 & \cellcolor[HTML]{F3F5F6} 1.04 & \cellcolor[HTML]{F0F4F6} 1.94 & \cellcolor[HTML]{ECF2F5} 2.99 & \cellcolor[HTML]{F9F0EB} -2.53 & \cellcolor[HTML]{FCE0D0} -7.90 & \cellcolor[HTML]{E3EDF3} 5.39 & \cellcolor[HTML]{F8F3F0} -1.51 & \cellcolor[HTML]{DEEBF2} 6.57 & \cellcolor[HTML]{F8F1ED} -2.32 \\
tiger shark & \cellcolor[HTML]{F9EFE9} -3.10 & \cellcolor[HTML]{E3EDF3} 5.41 & \cellcolor[HTML]{ECF2F5} 2.78 & \cellcolor[HTML]{DDEBF2} 6.82 & \cellcolor[HTML]{DEEBF2} 6.43 & \cellcolor[HTML]{FDDCC9} -9.53 & \cellcolor[HTML]{FCD7C2} -10.57 & \cellcolor[HTML]{E9F0F4} 3.57 & \cellcolor[HTML]{F9EBE3} -4.25 & \cellcolor[HTML]{EDF2F5} 2.36 & \cellcolor[HTML]{93C6DE} 19.91 \\
toilet seat & \cellcolor[HTML]{EAF1F5} 3.39 & \cellcolor[HTML]{F3F5F6} 0.93 & \cellcolor[HTML]{EFF3F5} 2.32 & \cellcolor[HTML]{F2F5F6} 1.29 & \cellcolor[HTML]{E7F0F4} 4.10 & \cellcolor[HTML]{F0F4F6} 1.75 & \cellcolor[HTML]{F9EDE5} -3.85 & \cellcolor[HTML]{F2F5F6} 1.47 & \cellcolor[HTML]{F6F7F7} 0.33 & \cellcolor[HTML]{F8F1ED} -2.20 & \cellcolor[HTML]{E1EDF3} 5.56 \\
tree frog & \cellcolor[HTML]{F8F4F2} -1.06 & \cellcolor[HTML]{E4EEF4} 5.06 & \cellcolor[HTML]{FAE9DF} -4.86 & \cellcolor[HTML]{E7F0F4} 3.97 & \cellcolor[HTML]{EFF3F5} 2.04 & \cellcolor[HTML]{FCE0D0} -8.17 & \cellcolor[HTML]{FCE2D2} -7.44 & \cellcolor[HTML]{E3EDF3} 5.33 & \cellcolor[HTML]{F7F5F4} -0.54 & \cellcolor[HTML]{E1EDF3} 5.81 & \cellcolor[HTML]{D2E6F0} 9.46 \\
umbrella & \cellcolor[HTML]{F5F6F7} 0.47 & \cellcolor[HTML]{ECF2F5} 3.05 & \cellcolor[HTML]{F9F0EB} -2.56 & \cellcolor[HTML]{F0F4F6} 1.95 & \cellcolor[HTML]{E0ECF3} 6.10 & \cellcolor[HTML]{FAEAE1} -4.49 & \cellcolor[HTML]{FDD9C4} -10.43 & \cellcolor[HTML]{F5F6F7} 0.66 & \cellcolor[HTML]{F9F0EB} -2.60 & \cellcolor[HTML]{F9EDE5} -3.83 & \cellcolor[HTML]{F8F1ED} -2.13 \\
vase & \cellcolor[HTML]{FBE6DA} -6.07 & \cellcolor[HTML]{DDEBF2} 6.83 & \cellcolor[HTML]{F8F4F2} -1.16 & \cellcolor[HTML]{DEEBF2} 6.58 & \cellcolor[HTML]{F2F5F6} 1.47 & \cellcolor[HTML]{FDDBC7} -9.80 & \cellcolor[HTML]{E8896C}\color[HTML]{F1F1F1}-23.88 & \cellcolor[HTML]{E0ECF3} 5.99 & \cellcolor[HTML]{F8F2EF} -1.78 & \cellcolor[HTML]{E6EFF4} 4.32 & \cellcolor[HTML]{FDDCC9} -9.68 \\
water bottle & \cellcolor[HTML]{FAE8DE} -5.38 & \cellcolor[HTML]{EDF2F5} 2.42 & \cellcolor[HTML]{F9EBE3} -4.21 & \cellcolor[HTML]{F0F4F6} 1.72 & \cellcolor[HTML]{F7F5F4} -0.78 & \cellcolor[HTML]{FDD9C4} -10.23 & \cellcolor[HTML]{EA8E70}\color[HTML]{F1F1F1}-23.10 & \cellcolor[HTML]{E1EDF3} 5.84 & \cellcolor[HTML]{F9EDE5} -3.55 & \cellcolor[HTML]{EDF2F5} 2.40 & \cellcolor[HTML]{FAC8AF} -13.30 \\
water tower & \cellcolor[HTML]{FBE6DA} -6.21 & \cellcolor[HTML]{DAE9F2} 7.72 & \cellcolor[HTML]{F2F5F6} 1.21 & \cellcolor[HTML]{DEEBF2} 6.43 & \cellcolor[HTML]{EAF1F5} 3.46 & \cellcolor[HTML]{F19E7D} -20.76 & \cellcolor[HTML]{E27B62}\color[HTML]{F1F1F1}-26.11 & \cellcolor[HTML]{D4E6F1} 9.24 & \cellcolor[HTML]{FAE7DC} -5.83 & \cellcolor[HTML]{DBEAF2} 7.11 & \cellcolor[HTML]{DAE9F2} 7.51 \\
wild boar & \cellcolor[HTML]{F7F5F4} -0.40 & \cellcolor[HTML]{EFF3F5} 2.23 & \cellcolor[HTML]{E7F0F4} 4.13 & \cellcolor[HTML]{E6EFF4} 4.55 & \cellcolor[HTML]{E0ECF3} 6.02 & \cellcolor[HTML]{FCDFCF} -8.31 & \cellcolor[HTML]{FCD5BF} -11.07 & \cellcolor[HTML]{ECF2F5} 2.95 & \cellcolor[HTML]{F6F7F7} 0.07 & \cellcolor[HTML]{F5F6F7} 0.61 & \cellcolor[HTML]{A5CEE3} 16.84 \\
wood rabbit & \cellcolor[HTML]{F9EFE9} -2.93 & \cellcolor[HTML]{F0F4F6} 1.61 & \cellcolor[HTML]{F8F3F0} -1.19 & \cellcolor[HTML]{EAF1F5} 3.21 & \cellcolor[HTML]{E7F0F4} 4.11 & \cellcolor[HTML]{FAE9DF} -4.85 & \cellcolor[HTML]{FCDECD} -8.97 & \cellcolor[HTML]{F8F2EF} -1.94 & \cellcolor[HTML]{F8F2EF} -1.84 & \cellcolor[HTML]{F2F5F6} 1.35 & \cellcolor[HTML]{CAE1EE} 11.30 \\
yawl & \cellcolor[HTML]{FCD5BF} -11.32 & \cellcolor[HTML]{C2DDEC} 12.36 & \cellcolor[HTML]{E3EDF3} 5.23 & \cellcolor[HTML]{C0DCEB} 12.78 & \cellcolor[HTML]{D4E6F1} 9.26 & \cellcolor[HTML]{D6604D}\color[HTML]{F1F1F1}-29.96 & \cellcolor[HTML]{CC4C44}\color[HTML]{F1F1F1}-32.80 & \cellcolor[HTML]{B3D6E8} 14.67 & \cellcolor[HTML]{FBE6DA} -6.21 & \cellcolor[HTML]{CAE1EE} 11.11 & \cellcolor[HTML]{E9F0F4} 3.67 \\
\bottomrule
\end{tabular}
}
\end{table}

\begingroup
\nopagebreak[4]
\subsubsection{Findings.}
The following interesting background biases are directly visible:
    \begin{itemize}
        \item
        \textbf{Some backgrounds are generally difficult} for all concepts: \textsf{machinery} generally poses a difficult background category (possibly due to the rich texture), while \textsf{open lands}  and \textsf{field} seem to be easier to distinguish from foregrounds. Alarmingly, one of the \emph{categories consistently causing drops in segmentation quality are road scenes}.
        
        \item
        \textbf{There are concept-specific biases:} Animals like \textsf{red fox} (\iouresultstablesimagenet{}), and furniture like \textsf{grand piano} and \textsf{park bench} (\iouresultstablesimagenet{}) or \textsf{dining table} (\iouresultstables{}) have a boost in performance on \textsf{fields} and \textsf{open lands},
        but a \textbf{noticeable drop in detection accuracy on urban scenes like {\normalfont \textsf{road}} and {\normalfont \textsf{architecture}}}, which may pose a safety risk in certain applications.
        This is counterintuitive: Animals should better fit into vegetation, making a foreground-background differentiation more difficult
        (cf.\ the drop of \textsf{frog} and \textsf{hamster} for \textsf{botanical} backgrounds, \iouresultstablesimagenet{}).
        Thus, this most probably originates from a \textbf{Clever Hans effect}, i.e., the CE uses the background as additional evidence for the foreground class, even though it is unrelated.
        
        \item
        \textbf{Many (though not all) biases are intuitive}: CEs for indoor pets like \textsf{cat} (\iouresultstables{}) or \textsf{hamster} (\iouresultstablesimagenet{}) get a boost on indoor scenes, while wild animals and cattle (e.g., \textsf{fox}, \textsf{boar}, \textsf{elephant} in \iouresultstablesimagenet{}; \textsf{horse}, \textsf{cow}, \textsf{sheep} in \iouresultstablesvoc{}) and vehicles see---partly severe---drops in performance.
        \\However, some clearly urban concepts like \textsf{street signs} and \textsf{streetcar} (\iouresultstablesimagenet{}) or \textsf{trains} (\iouresultstablesvoc{}) also drop clearly on \textsf{road} and \textsf{architecture} scenes, which cannot be fully explained by label noise.
        This shows that \textbf{some biases can hardly be anticipated}, and the ones uncovered here require further investigation.
    \end{itemize}
\endgroup

\subsection{Background Bias in Concept Representations}

To answer the question, whether a change in background distribution also changes the global(ized) concept representation, 
we train both new LoCEs and Net2Vec CEs for each of the 3 background randomizations in order to compare them to standard CEs with no background randomization.

\subsubsection{Ablation Study.}

We first conducted an ablation study with respect to the benefit of the number of layers and background variants per foreground image.
Naively, one could assume that both layer selection and high number of background variants are vital to capture the full spectrum of effects and performance / similarity variance.
However, this would make CE validation difficult in practice, because considering many layers and background variants heavily increases testing effort.
Fortunately, our findings do not confirm the naive assumption:
\begin{description}

\item[Number of variants per background:]
In the global and globalized cases, the number of backgrounds is not crucial. Increasing the number of background variations for the given foregrounds has no---initially even a slight adverse---effect on IoU (see \autoref{fig:iou-num-bgs}).
This emphasizes that the considered C-XAI methods are few-shot analysis methods.

\item[Layers:]
The effects of background randomization are \textbf{very similar across layers}. The only notable difference is, as expected for more complex object concepts, that early layers have consistently lower IoU values than the later ones.
Results are summarized in \autoref{tab:iou-per-layer}.
This means that for analysis of background bias it should be sufficient to stick with a single later layer, thus substantially reducing the cost of CE training.

\begin{table}[tb]
    \centering
     \caption{Average IoU performance of global and globalized CEs per layer depth. Results are averaged over concepts and models, best per row marked \textbf{bold}. Note that GloCEs outperform global ones consistently.}
    \label{tab:iou-per-layer}
    \sisetup{table-format=1.2}
    \robustify\bfseries
    \begin{tabular}{l @{~~~}S@{$\pm$}S@{~~~} S@{$\pm$}S@{~~~} S@{$\pm$}S}
    & \multicolumn{2}{c}{early} & \multicolumn{2}{c}{middle} & \multicolumn{2}{c}{late}\\\midrule[\heavyrulewidth]
    GloCE   & 0.270427 & 0.263640 & 0.403006 & 0.265799 & \bfseries 0.475616 & 0.246728 \\
    Net2Vec & 0.226347 & 0.234791 & 0.379059 & 0.261700 & \bfseries 0.445968 & 0.260190 \\
    \bottomrule
    \end{tabular}
\end{table}

\item[Models:]
The same holds for model architectures:
Both cosine similarities (cf.\ standard deviation in \autoref{fig:cos-sim-models}) and IoU differences (cf.\ standard deviation in \autoref{fig:iou-test-vanilla-vs-rand}) indicate similar trends across models. This is pretty much irrespective of their architecture, training task, and dataset. Expectedly, larger object detector models with more complex tasks tend to exhibit richer latent space semantics, i.e., higher IoU scores.

\begin{figure}[tb]
    \centering
    \vspace*{-.5\baselineskip}
    \includegraphics[width=0.8\linewidth]{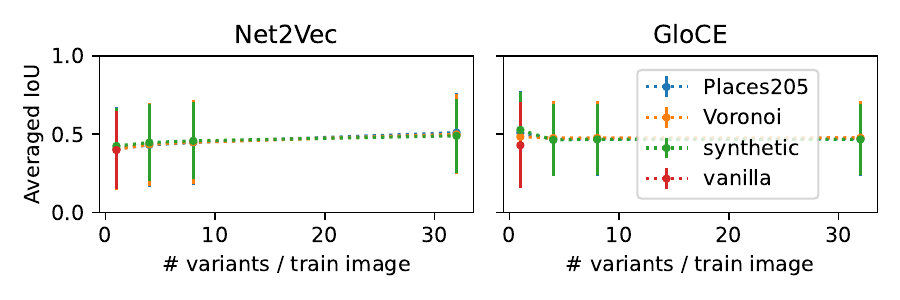}
    \vspace*{-0.75\baselineskip}
    \caption{Averaged IoU of CEs for different background randomization techniques at increasing amounts of training image variants. Variants are created by pasting a train image's foreground onto different backgrounds, using a plain \emph{Places205} background, a \emph{Voronoi} version, or a \emph{synthetic} background. The \emph{vanilla} baseline (single variant, no randomization) is marked in red.
    The mean is taken over concepts from ImageNetS50 and Places205, for each the late layer of 7 models.
    }
    \label{fig:iou-num-bgs}
\end{figure}

\item[CE Method:]
Net2Vec turned out to yield slightly worse IoU results at the same amount of training time (cf.\ \autoref{tab:iou-per-layer}). Also, GloCEs show slightly stronger relative deviation between vanilla and randomized trained versions (cf.\ the stronger coloring in \autoref{tab:iou-per-bg-cat-voc}), attesting them slightly better bias-capturing capabilities.
Improving convergence, e.g., by increasing the number of epochs, did not change above tendencies.
\end{description}
This means, \textbf{already a minimal setup of a single variant per foreground and a single layer per model can provide valuable insights into the robustness and bias of CEs.}

\subsubsection{Cosine Similarities.}

\begin{figure}[tbh]
        \centering
        \vspace*{-0.5\baselineskip}
        \includegraphics[width=\linewidth]{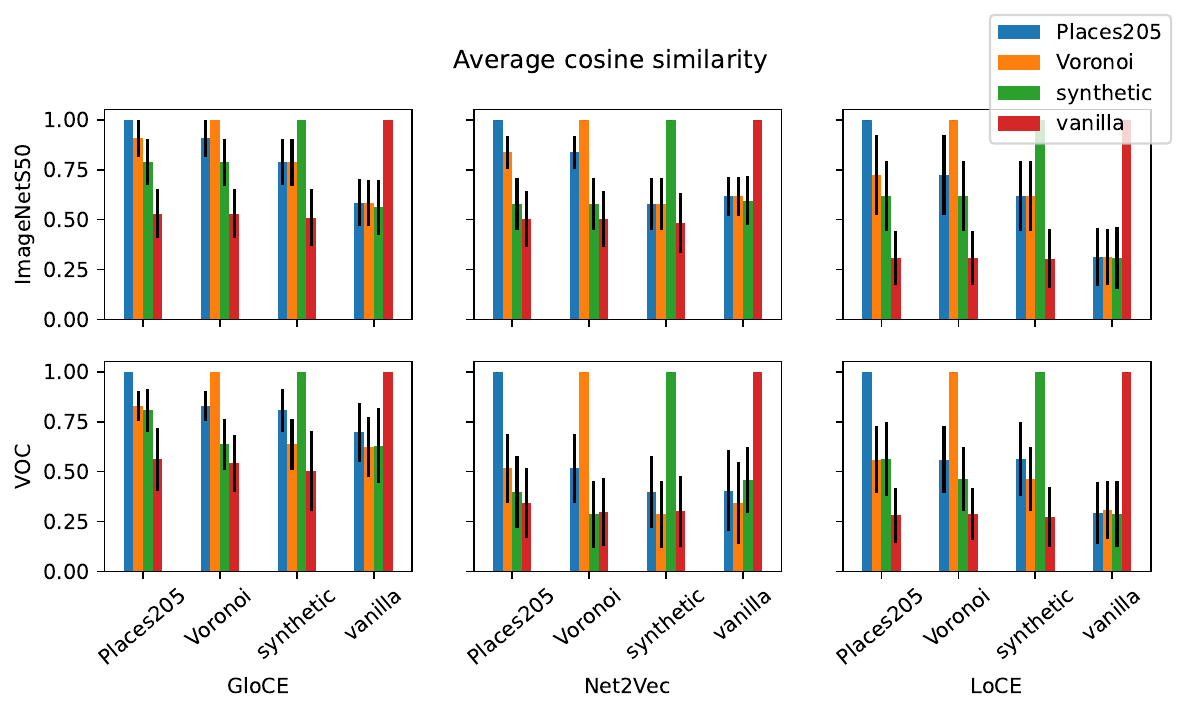}
        \vspace*{-\baselineskip}
        \caption{Pairwise cosine similarities between CEs of the same concept and layer but from different train data randomization schemes.
            Cosine similarities are calculated for matching pairs of CEs for the same layer (late layers only) and concept, then averaged. 
            Bars indicate left to right averaged similarity to CEs trained on \textit{Places205} (\textit{blue}), \textit{Voronoi} (\textit{orange}), \textit{synthetic} (\textit{green}), \textit{vanilla} (\textit{red}) data.
            Standard deviation is indicated via error bars (black). For visual reference, cosine similarities of CEs with themselves are reported (value of 1).}
            \vspace*{-\baselineskip}
        \label{fig:cos-sim-models}
    \end{figure}

The pairwise cosine similarities show for all CE methods that significantly different vectors are learned, as shown in \autoref{fig:cos-sim-models}. Interestingly, this is the case pairwise between \emph{all} available methods, only with a slight trend of stronger similarity between the two Places205 based techniques.
The dissimilarity between these two underscores the \textbf{non-negligible effect of using shape randomization for concept segmentation}.

Finally, local methods yield much more dissimilar vectors for the different techniques. However, this is mostly averaged out when globalizing, i.e., averaging, confirming previous results \cite{mikriukov2024local}, that \textbf{local overfitting of LoCEs well captures the full variance of a concept representation}. Also, when using cosine similarity as a measure of stability, \textbf{local-to-global methods seem to overtake purely global ones} in demanding settings, like this one with large background variance.

\subsection{Performance Changes}

\begin{figure}[tb]
    \centering
    \includegraphics[width=0.75\linewidth]{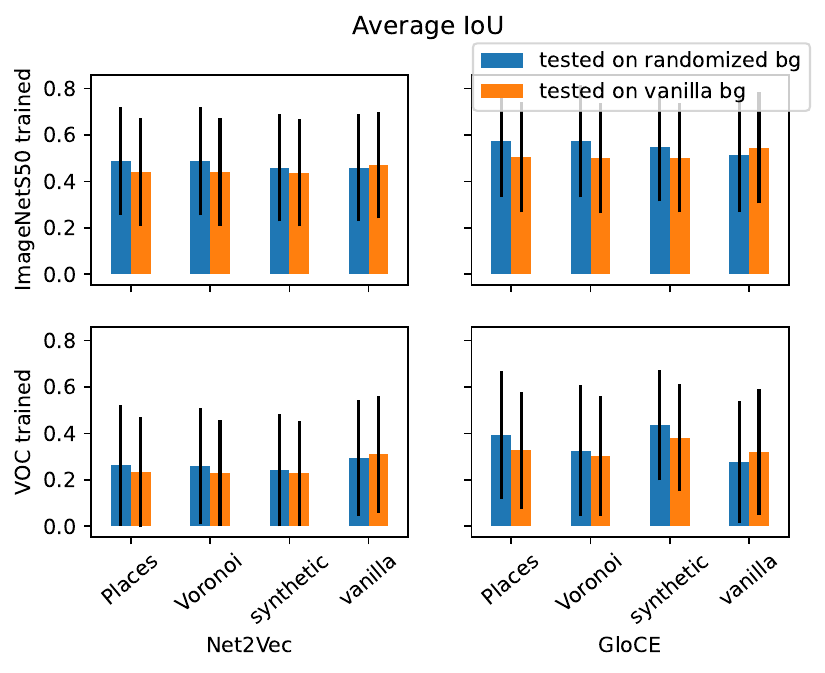}
    \\[-\baselineskip]
    \caption{IoU on bg-randomized test images (simple bg pasting from Places validation dataset) versus non-randomized ones.
    Compared are different training schemes: Pasting the test image foregrounds onto single unchanged or \emph{Voronoi} backgrounds from the \emph{Places} training data, onto \emph{synthetic} backgrounds, or keeping the original backgrounds (\emph{vanilla}).
    Results are shown for Net2Vec (\emph{left}) and GloCE (\emph{right}) CEs trained for concepts from Pascal VOC (\emph{top}) and ImageNetS50 (\emph{bottom}), averaged over all concepts and 7 DNNs.}
    \label{fig:iou-test-vanilla-vs-rand}
\end{figure}

As to be expected, both global methods receive a boost in generalization to background-randomized data if trained on such. Interestingly, they simultaneously show a comparatively small to no loss in performance on vanilla test data,
maintaining the level of vanilla-trained models; and thus \textbf{CEs with background randomized training data outperform the vanilla ones clearly on mixed test datasets}.
This holds throughout all considered layer depths, most clearly visible in early layers. Results on late layers are shown in \autoref{fig:iou-test-vanilla-vs-rand}.
Interestingly, also the more sophisticated Voronoi approach, which brings in way more background information per image, is no clear winner with respect to performance.
This is good news for scalability: The simple randomization techniques employed here already focused on a cost-effective setup. If already such a simple and easy-to-employ technique can reveal interesting insights, these could serve many practical use-cases instantly.

Lastly, we revisit the per-background-class testing setup in \autoref{sec:exp-testing}.
The CEs arising from background-randomized training can be understood as a baseline for the performance achievable using background-agnostic training. While the previous comparison rather revealed consistently under- or overperforming background categories (column-wise results), here we find consistency on concept side (row-wise results): Concepts like
\textsf{cow} (\iouresultstablesvoc{}); \textsf{dog}, \textsf{dining table} (\iouresultstables{}); and \textsf{motorbike} and \textsf{person} (\iouresultstablesimagenet{})
consistently show major improvements when using background randomization during training. This indicates that (1) these concepts probably had an issue with background bias in the concept training data; and 
(2)  \textbf{solely \textit{testing} on background randomized data will not reveal all flaws related to background bias, but background-randomized CEs can help to add these insights.}
  
\subsection{Discussion and Future Work}

\paragraph{Limitations: Label Noise.}
Our real-image background dataset does not feature segmentation labels for the concepts. Thus, we reside to filtering out background images that could possibly contain the concept of interest, which would invalidate the original concept mask. This, however, is error prone, and might change the distribution of the background images in a category (e.g., most \textsf{indoor} scenes contain a \textsf{chair}, cf.\ \iouresultstablesvoc{}).
We obtained the same trends, just generally lower IoU values, without the filtering, suggesting that the distribution change does not invalidate results per se.
Nevertheless, practical application could consider scene datasets with segmentation labels for the concepts instead.

\paragraph{Additional Models and Datasets.}
It should be noted that clearly the setup could be widened even further, e.g., including more background datasets to investigate biases with respect to backgrounds not part of Places205 or our synthetic background classes, and in other domains like medical diagnostics. Also, checking results on most recent models, like later YOLO and DETR versions, would indicate whether next generation architectures show different trends. However, with >50 concepts from two quite different datasets, and 7 diverse state-of-the-art models, we are confident that the results generalize to most standard models and yield also interesting results for further background classes.

\paragraph{Distribution Perspective.}
Some interesting future work would be to further leverage the distribution nature of LoCEs: Our results suggest that they encode a lot of valuable information about variance and concept corner cases.
Hence, instead of sole cosine similarity between vectors, one could compare the distributions. For this, different techniques to model the distribution out of a set of samples could be investigated: Apart from the Gaussian mixture modeling in \cite{mikriukov2024local}, also simpler techniques like PCA might already be sufficient.
Comparison of distributions instead of vectors would then hopefully give some hints on typical indicators for flawed concept representations in latent spaces; eventually opening doors to fix them in a post-hoc targeted manner, or incorporate findings about skewed distributions into the training objective.

\paragraph{Downstream Tasks: Model Fixing.}
Finally, a valuable future direction would be to further assess the actual practical impact of background-biased concept embeddings: Do the identified background biases allow the construction of DNN failure cases? And how useful is this information to fix the DNN? These questions need to be answered on the way to practical application of this method.


\section{Conclusion}

This paper revisited concept-based XAI techniques for post-hoc concept segmentation.
Our guiding question, namely, whether state-of-the-art techniques exhibit a background bias,
was finally answered affirmative:
Across models, layers, concepts, and datasets we were able to identify notable drops in performance of concept segmentation models when changing the background distributions.
E.g., wild animals being less well detected on roads.
Gladly, we were able to show that any such flaws can be dug out with relatively cheap and simple background randomization techniques, as presented in this work.
This hopefully motivates readers to introduce beneficial background randomization techniques also into other data-driven explainability techniques; and finally eases investigation of DNN biases encoded in their latent spaces.

\begingroup
\small
\section*{Acknowledgments}
G.S. acknowledges support through the junior research group project ``chAI'' funded by the German Federal Ministry of Education and Research (BMBF), grant no.\ 01IS24058. The authors are solely responsible for the content of this publication.
We acknowledge financial support by Land Schleswig-Holstein within the funding program Open Access Publikationsfund.
E.H., A.M.\ and M.R.\ acknowledge support through the junior research group project ``UnrEAL'' by the German Federal Ministry of Education and Research (BMBF), grant no.\ 01IS22069.
S.G. acknowledges support by a studentship from the School of Electrical Engineering, Electronics and Computer Science, at the University of Liverpool, UK.
\endgroup


%
%
\bibliographystyle{splncs04}
\bibliography{main}

\end{document}